\documentclass[10pt,twocolumn,letterpaper]{article}

\usepackage[pagenumbers]{cvpr} %

\usepackage{graphicx}
\usepackage{amsmath}
\usepackage{amssymb}
\usepackage{booktabs}
\usepackage{caption}
\usepackage{cuted}
\usepackage{makecell}
\usepackage{xcolor}
\usepackage{pifont}
\usepackage{cite}
\usepackage{float}
\usepackage[resetlabels]{multibib}

\usepackage[accsupp]{axessibility}  %

\usepackage[pagebackref,breaklinks,colorlinks]{hyperref}

\usepackage[capitalize]{cleveref}
\crefname{section}{Sec.}{Secs.}
\Crefname{section}{Section}{Sections}
\Crefname{table}{Table}{Tables}
\crefname{table}{Tab.}{Tabs.}

\makeatletter
\renewcommand{\paragraph}{%
  \@startsection{paragraph}{4}%
  {\z@}{.5ex \@plus 1ex \@minus .2ex}{-1em}%
  {\normalfont\normalsize\bfseries}%
}
\makeatother

\def\cvprPaperID{11845} %
\def\confName{CVPR}
\def\confYear{2022}

\begin{document}

\title{
    Deep Spectral Methods: A Surprisingly Strong Baseline for Unsupervised Semantic Segmentation and Localization
}

\author{
Luke Melas-Kyriazi\\
\and
Christian Rupprecht\\
\and
Iro Laina\\
\and
Andrea Vedaldi\\
\and 
University of Oxford \\
{\tt\small \{lukemk,chrisr,iro,vedaldi\}@robots.ox.ac.uk}
}
\maketitle

\begin{abstract}
Unsupervised localization and segmentation are long-standing computer vision challenges that involve decomposing an image into semantically meaningful segments without any labeled data.
These tasks are particularly interesting in an unsupervised setting due to the difficulty and cost of obtaining dense image annotations, but existing unsupervised approaches struggle with complex scenes containing multiple objects. 
Differently from existing methods, which are purely based on deep learning, we take inspiration from traditional spectral segmentation methods by reframing image decomposition as a graph partitioning problem. 
Specifically, we examine the eigenvectors of the Laplacian of a feature affinity matrix from self-supervised networks.
We find that these eigenvectors already decompose an image into meaningful segments, and can be readily used to localize objects in a scene.
Furthermore, by clustering the features associated with these segments across a dataset, we can obtain well-delineated, nameable regions, i.e.\ semantic segmentations.
Experiments on complex datasets (PASCAL VOC, MS-COCO) demonstrate that our simple spectral method outperforms the state-of-the-art in unsupervised localization and segmentation by a significant margin.
Furthermore, our method can be readily used for a variety of complex image editing tasks, such as background removal and compositing. 
\footnote{Project Page: \href{https://lukemelas.github.io/deep-spectral-segmentation/}{https://lukemelas.github.io/deep-spectral-segmentation/}}
\end{abstract}

\section{Introduction}
\begin{figure}
  \centering
  \includegraphics[width=0.99\linewidth]{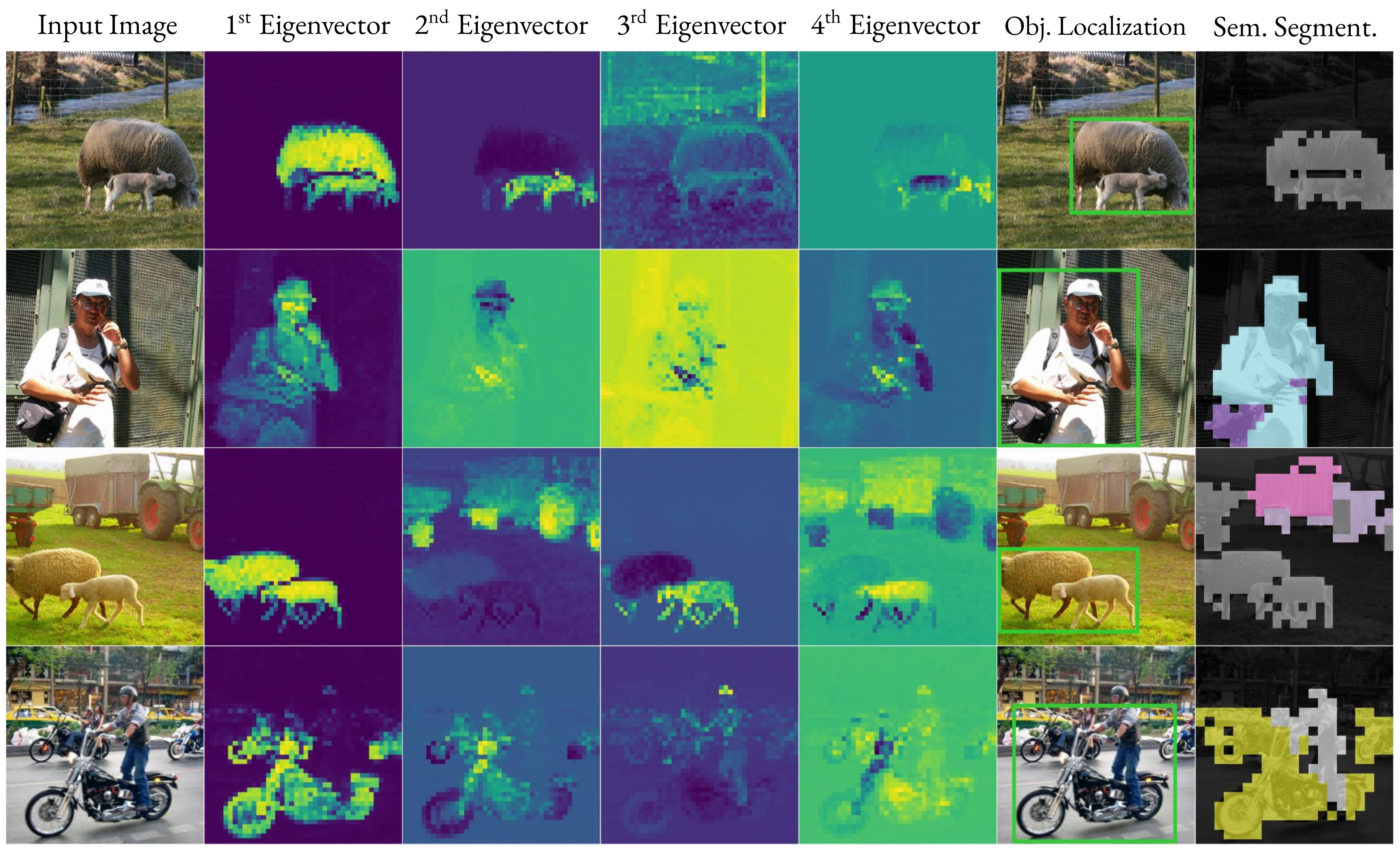}
  \caption{\textbf{Deep Spectral Methods.}
    We present a simple approach based on spectral methods that decomposes an image using the eigenvectors of a Laplacian matrix constructed from a combination of color information and unsupervised deep features. The method surpasses the state of the art in unsupervised image segmentation and object localization while also being significantly simpler.
  }\label{fig:splash}
\end{figure}

Well-established computer vision tasks such as localization and segmentation are aimed at understanding the structure of images at a fine level of detail. 
The modern approach to these tasks, which consists of training deep neural networks in an end-to-end fashion, has shown strong performance given large quantities of human-labeled data. 
However, obtaining labeled data for these dense tasks can be difficult and expensive. 
Whereas vast quantities of (weak) image labels and descriptions may be obtained from the Internet \cite{radford2021learning,mahajan2018exploring,jia2021scaling}, dense image annotations cannot be easily sourced from it, and creating them manually is a labor-intensive process.
Moreover, in many specialized fields such as medical imaging, where detection and segmentation tasks are particularly important, data labeling must be manually performed by a domain expert. 
As a result, performing dense vision tasks without labeled data is an important open problem.

Numerous existing methods perform dense visual tasks with weak annotations, such as image-level labels, captions, spoken narratives, scribbles, and points \cite{khoreva_CVPR17,huang2018deep,chan2019comprehensive,ke2021spml}.
However, fully-unsupervised dense image understanding remains challenging and under-explored. 
Current approaches involve clustering dense features (\eg, IIC~\cite{ji2018invariant}), contrastive learning with saliency masks (\eg, MaskContrast~\cite{vangansbeke2020unsupervised}), and 
GAN-based object discovery (\eg, ReDo~\cite{chen_unsupervised_2019}). 
Others segment images of a given object category into a number of semantic regions (parts)~\cite{hung2019scops,choudhury2021unsupervised,collins2018deep}.
However, these approaches tend to struggle with complex images, and most of them can only identify a single object per image.

In this paper, we take inspiration from image segmentation methods from the pre-deep learning era, which framed the segmentation problem as one of graph partitioning.
Whereas existing unsupervised segmentation methods are based primarily on deep learning, we show the benefits of combining deep learning with traditional graph-theoretic methods. 

Our method first utilizes a self-supervised network to extract dense features corresponding to image patches. 
We then construct a weighted graph over patches, where edge weights give the semantic affinity of pairs of patches, and we consider the eigendecomposition of this graph’s Laplacian matrix.
We find that without imposing any additional structure, the eigenvectors of the Laplacian of this graph directly correspond to semantically meaningful image regions. 
Notably, the eigenvector with the smallest nonzero eigenvalue generally corresponds to the most prominent object in the scene. 
We show that, surprisingly, simply extracting bounding boxes or masks from this eigenvector surpasses the current state of the art on unsupervised object localization/segmentation across numerous benchmarks. %

Next, we propose a pipeline for semantic segmentation. 
We first convert the eigensegments into discrete image regions by thresholding and associate each region with a semantic feature vector from the network. 
We consider all image regions in a large dataset of images and jointly cluster these regions, yielding semantic (pseudo-)labels that are consistent across the dataset. 
Lastly, we perform self-training using these labels to refine our results, and we evaluate against the ground truth segmentations. 
Different from prior self-supervised semantic segmentation methods, our method performs well on complex images without fine-tuning. 
Importantly, while recent GAN-based and saliency-based methods are limited to finding a single semantic region per image,  our method can segment multiple semantic regions and outperforms prior methods on PASCAL VOC 2012, re-surfacing spectral methods as a strong baseline for future work. 

Finally, we show that a slight variant of our method is well-suited to the task of soft image decomposition (\ie, image matting), or breaking down an (RGB) image into multiple (RGB-A) layers with soft boundaries. 
This decomposition dramatically simplifies real-world image editing and compositing tasks such as background replacement. 

\section{Related work}\label{s:related}

In this section, we discuss the different fields related to our approach along with the relatively new tasks of unsupervised localization and segmentation.

\paragraph{Self-Supervised Visual Representation Learning}

The past five years have seen tremendous progress in self-supervised visual representation learning. Early research in this area was based on solving pretext tasks such as rotation, jigsaw puzzles, colorization, and inpainting \cite{doersch2015unsupervised,noroozi2016unsupervised,pathak2016context,oord2018representation,hnaff2019dataefficient,zhang2016colorful,zhang2017split,gidaris2018unsupervised}. 
Recent methods mostly consist of contrastive learning with heavy data augmentation~\cite{gutmann2010noise,oord2018representation,wu2018unsupervised,chen2020simple,kalantidis2020hard,he2019moco,chen2020improved,henaff19data-efficient,tian19contrastive,tian2020makes}. 
Others distinguish themselves by removing the dependency on negative examples~\cite{zbontar2021barlow,chen2021exploring,grill2020byol}, the use of clustering~\cite{caron2020unsupervised,li2021prototypical}, and most recently, the extension to transformer architectures~\cite{caron2021emerging,chen2021mocov3}.

As these methods primarily focused on the downstream task of image classification, they were designed to produce a single global feature vector for each input image. 
Although numerous specialized methods have been proposed for dense contrastive learning \cite{xie2020propagate,wang2020DenseCL,pinheiro2020unsupervised}, they require fine-tuning for application on tasks such as detection or segmentation. 
Recently, however, the emergence of self-supervised vision transformers has made it possible to extract dense (patch-wise) feature vectors without the need for specialized dense contrastive learning methods.

\paragraph{Vision Transformers}

The past year in computer vision research has been marked by the emergence of the vision transformer (ViT) architecture, a variant of the transformer model popular in sequence modeling and NLP \cite{vaswani17attention}. 
Vision transformers reshape an image into a sequence of small patches 
and apply layers of self-attention before aggregating the result into an additional global token, usually denoted \texttt{[CLS]}. 
Numerous variants of the vision transformer have been proposed, such as Deit~\cite{touvron2020deit}, Token-to-Token ViT~\cite{yuan2021tokens}, DeepViT~\cite{zhou2021deepvit}, CrossViT~\cite{chen2021crossvit}, PiT~\cite{heo2021rethinking}, CaiT~\cite{touvron2021cait}, LeViT~\cite{graham2021levit}, CvT~\cite{wu2021cvt}, and Swin/Twins~\cite{liu2021swin,chu2021twins}. 
Nonetheless, the defining aspect of this class of models is the use of self-attention to process information throughout the network. 
Since self-attention involves self-comparisons of image patch features, it is natural to construct a semantic affinity matrix over patches, as we do in our approach. 

Numerous recent works \cite{chen2021mocov3,caron2021emerging,atito2021sit} have observed that vision transformers perform exceptionally well relative to convolutional architectures in the self-supervised setting. 
In particular, Caron \etal~\cite{caron2021emerging} train a self-supervised ViT using multi-crop training and a momentum encoder, and argue that it learns better-localized features than supervised ViTs or ResNets. 
They show that the self-attention layer of the last \texttt{[CLS]} token has heads that localize foreground objects in the scene. 
This is a baseline in our object localization experiments, upon which we improve substantially.

\paragraph{Unsupervised Localization}

Unsupervised localization refers to the task of finding the location of an object in the form of a bounding box. 
Most approaches to localization find objects by identifying co-occurring patterns across image collections \cite{cho2015unsupervised,vo2019unsupervised,vo2020toward}. However, due to inter-image comparisons, these approaches are generally slow, memory-intensive, and have trouble scaling to large datasets. 

Recently, some works have begun to eschew this paradigm by finding objects purely from the features of pretrained deep networks. 
\cite{collins2018deep} perform co-localization through deep feature factorization.
\cite{zhang2020object} mines patterns from pretrained convolutional network features. 
Most recently, LOST~\cite{LOST} extracts localization information from the attention features of a self-supervised vision transformer. 
LOST uses the same features as we do in our approach, but extracts bounding boxes using seed selection and expansion based on the number of patch correlations.
By contrast, our approach localizes objects through spectral bipartitioning, which is more principled and more performant than \cite{LOST}. 

\paragraph{Unsupervised Segmentation}

Whereas a plethora of methods have tackled segmentation in semi- and weakly-supervised settings~\cite{khoreva_CVPR17,huang2018deep,chan2019comprehensive,ke2021spml}, the unsupervised setting remains a relatively nascent area of research. 
Current methods can broadly be characterized as either generative or discriminative. Generative methods use unlabeled images to train special-purpose image generators \cite{chen_unsupervised_2019,arandjelovic2019object,abdal2021labels4free}, or directly extract segmentations from pretrained generators \cite{voynov2021object,melas2021finding}; 
discriminative methods are primarily based on clustering and contrastive learning ~\cite{ji2018invariant,hwang2019segsort,zhang2020self,vangansbeke2020unsupervised}. 
MaskContrast~\cite{vangansbeke2020unsupervised}, the current state of the art in unsupervised semantic segmentation, uses saliency detection to find object segments (\ie, the foreground) and then learns pixel-wise embeddings via a contrastive objective. 
However, MaskContrast relies heavily on a saliency network which is initialized with a pretrained (fully-supervised) network. 
Moreover, it heavily relies on the assumption that all foreground pixels in a given image belong to the same object category, which fails to hold for most images containing multiple objects. 

\paragraph{Spectral Methods}
Spectral graph theory emerged from the study of continuous operators on Riemannian manifolds~\cite{cheeger1969alower}. 
Subsequent works brought this line of research to the discrete setting of graphs, with numerous results connecting global properties of graphs to the eigenvalues and eigenvectors of their Laplacian matrices. %
Two of these results are essential for understanding our work. 
First, \cite{fiedler1973algebraic} showed that the second-smallest eigenvalue of a graph, now called the algebraic connectivity or the Fiedler eigenvalue, quantifies the connectivity of a graph. 
In our work, we use the Fiedler eigenvector for object localization. 
Second, \cite{fiedler1973algebraic,donath1973lower} showed that the eigenvectors of graph Laplacians yield minimum-energy graph partitions. 
In our work, we use this idea to extract semantic segmentations from a patch-wise \emph{semantic} affinity matrix. 

Spectral graph methods were popularized within the machine learning and computer vision communities by \cite{shi2000normalized, ng01on-spectral}. 
Shi and Malik~\cite{shi2000normalized} framed image segmentation in the language of graph cuts, that is finding a partition of the graph (\ie the image) to minimize the similarity of the partitions. 
They noted that finding the minimum cut generally leads to smaller-than-desired partitions, so they normalized by the total weight of all edges connected to each partition. 
Ng \etal~\cite{ng01on-spectral} also performed spectral decomposition, 
and then stacked and clustered the eigenvectors along the eigenvector dimension in order to obtain a fixed number of partitions.
Numerous follow-up works have extended these ideas to new settings such as co-segmentation \cite{wu2013unsupervised}. 
\begin{figure*}
  \centering
  \includegraphics[width=0.93\textwidth]{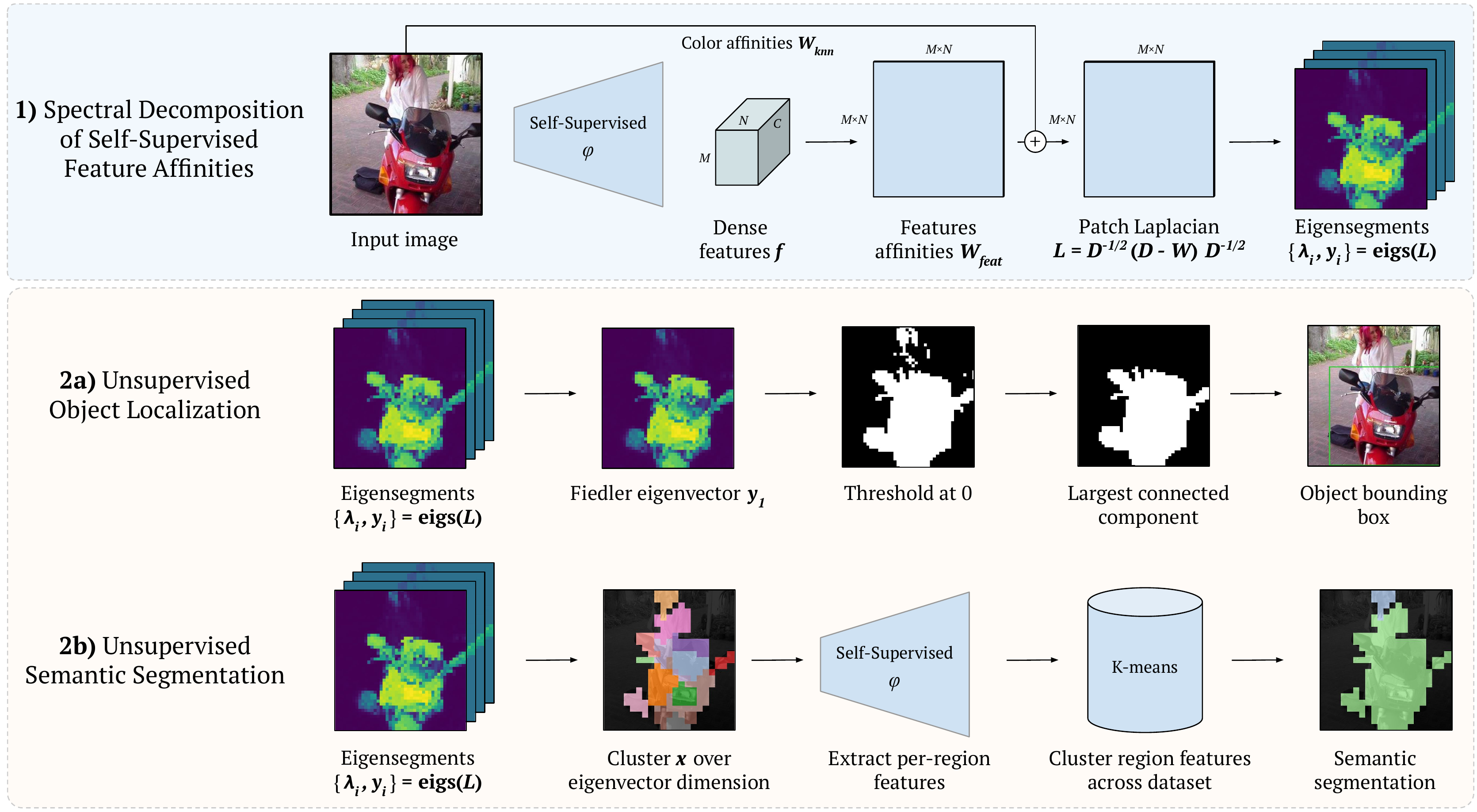}
  \caption{\textbf{Method overview.}
    Our approach to unsupervised segmentation combines the benefits of modern self-supervised learning with the benefits of traditional spectral methods. Given an image, we first extract dense features from a network $\phi$ and use these to construct a semantic affinity matrix, which we then fuse with low-level color information. We decompose the image into soft segments by computing the eigenvectors of the Laplacian of this matrix. Second, we can use these eigenvectors for a wide range of downstream tasks. For object localization (2a), we find that simply taking the sign of the eigenvector with the smallest nonzero eigenvalue, and placing a bounding box around this region, produces state-of-the-art object localization performance. For semantic segmentation (2b), we convert the eigenvectors into discrete segments, compute a feature vector for each segment, and cluster these segments across an entire dataset.
  }\label{fig:method_diagram}
\end{figure*}

\section{Method}\label{s:method}

Our method is rooted in ideas from spectral graph theory, the study of analyzing properties of graphs by looking at their spectra (\ie their eigenvalues).
We begin with a general overview of spectral methods.

\subsection{Background}\label{s:background}

Let $G = (V, E)$ be a weighted undirected graph with adjacency matrix $W = \{ w(u, v) : (u, v) \in E \}$.
The Laplacian matrix $L$ of this graph is given by $L = D - W$ or in the normalized case $L = D^{-1/2}(D - W)D^{-1/2}$, where $D$ is the diagonal matrix whose entries contain the row-wise sum of $W$.  
The Laplacian is particularly interesting because it corresponds to a quadratic form 
\[ x^T L x = \sum_{(u,v) \in E} w(u,v) \cdot (x(u) - x(v))^{2} \]
for $x \in \mathbb{R}^{n}$ with $n = |V|$. 
Intuitively, this quadratic form measures the smoothness of a function $x$ defined on the graph $G$. If $x$ is smooth with respect to $G$, then $x(u)$ is similar to $x(v)$ whenever $u$ is similar to $v$ (where similarity is quantified by the weight $w(u,v)$). 

The eigenvectors and eigenvalues of $L$ are the central objects of study in spectral graph theory. The eigenvectors $y_{i}$ span an orthogonal basis for functions on $G$ that is, in the sense described above, the \textit{smoothest} possible orthogonal basis: 
\[ y_{i} = \text{argmin}_{\|x\|=1, x \perp y_{<i}} x^T L x  \]
with $y_0 = \mathbf{1} \in \mathbb{R}^{n}$. 
The eigenvalues $\lambda_{i}$, with $0=\lambda_{0} \leq \lambda_{1} \leq \dots \leq \lambda_{n-1}$, are the values of the right-hand side of the above expression (which is known as the Courant-Fischer theorem in the spectral graph theory community). 
Thus, it is natural to express functions on $G$ in the basis of the eigenvectors of the graph Laplacian. 

In the context of classical image segmentation, the nodes of $G$ correspond to the pixels of an image $I\!\in\!\mathbb{R}^{MN}$, and the edge weights $W\!\in\!\mathbb{R}^{MN \times MN}$ correspond to the affinities between pairs of pixels. 
The eigenvectors $y \in \mathbb{R}^{MN}$ can be thought of as soft image segments. 
Discrete graph partitions are often called graph cuts, where the value of cut with two partitions $A$ and $B$ ($A \cup B = V$, $A \cap B = \emptyset$) 
is given by 
\[ 
\operatorname{cut}(A, B) = \sum_{u \in A, v \in B} w(u, v) 
\] 
\ie, the total weight of edges that have been removed by partitioning. 
A normalized version of graph cuts \cite{shi00normalized} is particularly useful in practice, and emerges naturally from the eigenvectors of the normalized Laplacian; the optimal (continuous) bi-partitioning in this case reduces to finding the Laplacian eigenvectors.

\paragraph{The choice of W}

The most important aspect of spectral methods for image segmentation is the construction of the adjacency matrix $W$, which encodes the similarities of each pair of pixels.

To take the spatial neighborhood and color information into account, some works (especially for image matting \cite{chen2013knn,lee2011nonlocal}) use a nearest neighbor formulation.
KNN-matting converts an image to the HSV color space, and defines for each pixel a vector $\psi(u) = (\cos(c_H), \sin(c_H), c_S, c_V, p_x, p_y) \in \mathbb{R}^5$ containing both color information (the $c_H, c_S, c_V$ values) and spatial information (the $p_x, p_y$ values). 
They then construct a sparse affinity matrix from pixel-wise nearest neighbors based on $\psi$:
\begin{equation} \label{eq:w_knn}
W_\mathrm{knn}(u, v) = \begin{cases}
    1 - \| \psi(u) - \psi(v) \|, & u \in \text{KNN}_{\psi}(v), \\
    0, & \text{otherwise},
\end{cases} 
\end{equation}
where $u \in \text{KNN}_{\psi}(v)$ are the $k$-nearest neighbors of $v$ under the distance defined by $\psi$. 

This affinity matrix typically yields sharp object boundaries, but since it is designed for the image matting setting, in which a trimap is given, it does not take semantic information into account.
While some recent works have explored the idea of semantic matting~\cite{sun2021sim,aksoy2018semanticsoft}, they tackle a different problem from us and are limited by their use of large labeled datasets. 
Our work combines the benefits of this classical spectral decomposition with the benefits of deep self-supervised features 
for the purpose of unsupervised object and scene understanding. 

\subsection{Semantic Spectral Decomposition}

In this section we will describe our formulation of deep spectral decomposition and its direct application to the down-stream tasks of object localization and semantic segmentation.
An overview is shown in \cref{fig:method_diagram}.

Let $I \in \mathbb{R}^{3 \times M \times N}$ be an image. 
We will first decompose $I$ into semantically consistent regions using features of a neural network. 
These regions can then be processed into bounding boxes for object localization or clustered across a collection of images for semantic segmentation. 

We begin by extracting deep features $f = \phi(I) \in \mathbb{R}^{C \times M/P \times N/P}$ using a network $\phi$. 
As the feature maps of a network are usually computed at a lower resolution than the image, we introduce $P$ for this downsampling factor.
In the case where $\phi$ is a transformer architecture, $P$ represents the patch size. 
These features may be extracted from any part of the network and can even be a combination of different layers, \ie, hyper-columns~\cite{hariharan2015hypercolumns}. 
In our experiments with transformers, we find, similarly to \cite{LOST}, that features from the keys of the last attention layer work especially well, as they are inherently meant for self-aggregation of similar features. 

We then construct an affinity matrix from the patchwise feature correlations. 
Additionally, we threshold the affinities at $0$, because the features are designed for aggregating similar features rather than anti-correlated features: 
\begin{equation}
    W_\mathrm{feat} = ff^{T} \odot (ff^{T} > 0)\in \mathbb{R}^{\frac{MN}{P^2} \times \frac{MN}{P^2}}
\end{equation}
These feature affinities contain rich semantic information at a coarse resolution. 
To gain back low-level details, we fuse them with traditional color-level information which can be seen as features from the 0-th layer of the network. 

To perform the fusion, we bilinearly upsample the features and downsample the image to an intermediate resolution $M'\times N'$. Empirically, we find that using $M' = M/8$ and $N' = N/8$ yields good low-level details while maintaining a fast runtime. 
As our color affinity matrix, we use the sparse KNN-matting matrix (\cref{eq:w_knn}), although any traditional similarity matrix can also be used.
Our final affinity matrix is the weighed sum of the feature and color matrices:
\begin{equation}
W = W_\mathrm{feat} + \lambda_\mathrm{knn} W_\mathrm{knn}
\end{equation}
where $\lambda_\mathrm{knn}$ is a user-defined parameter that trades off semantic and color consistency.

Given $W$, we take the eigenvectors of its Laplacian $L = D^{-1/2}(D - W)D^{-1/2}$ to decompose an image into soft segments: $\{y_0, \cdots, y_{n-1}\} = \mathrm{eigs}(L)$. 
Since the first eigenvector $y_0$, is a constant vector corresponding to $\lambda_0 = 0$, for our purposes we use only $y_{>0}$. 

\begin{figure}
  \centering
  \includegraphics[width=0.47\textwidth]{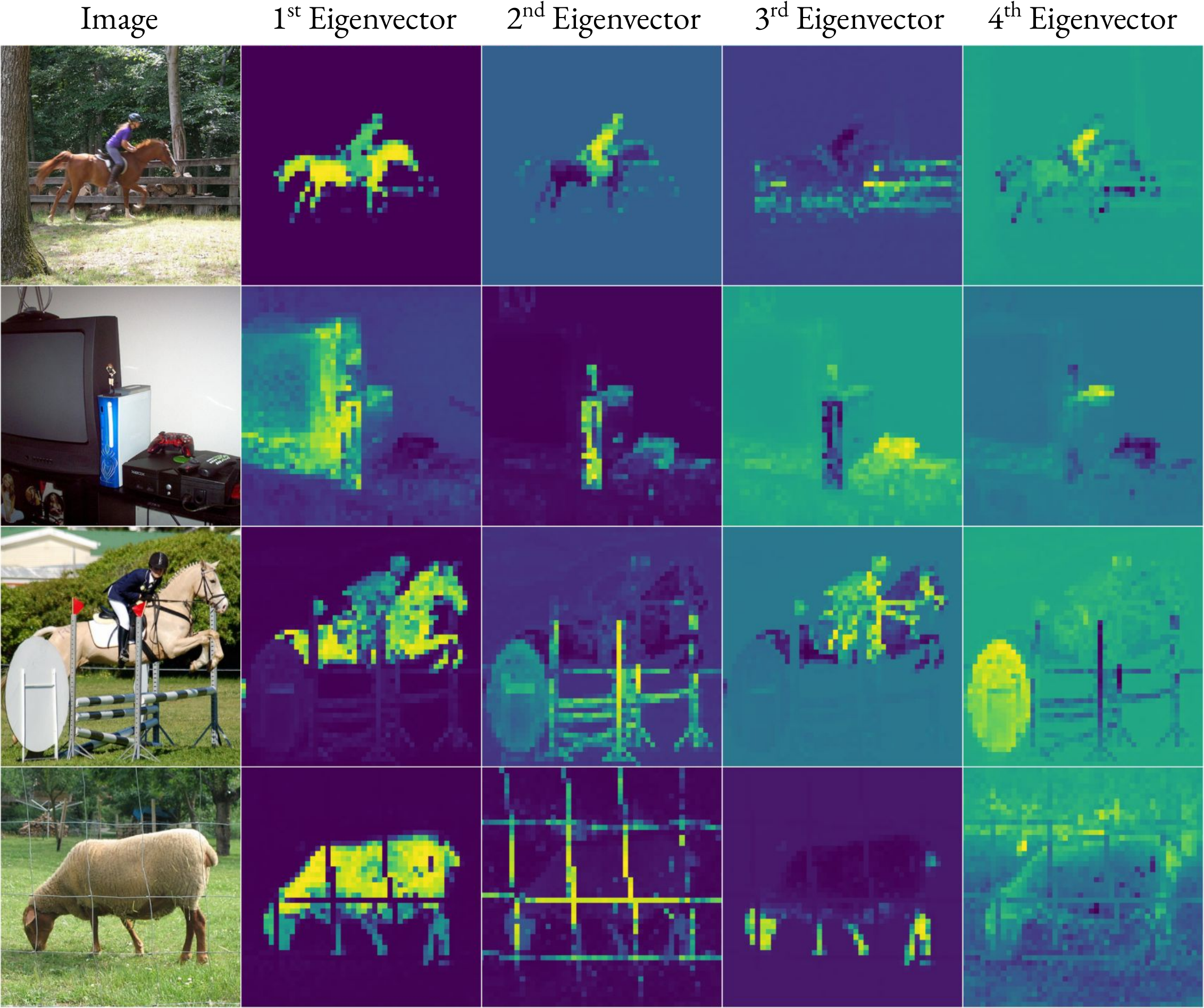}
  \caption{\textbf{Eigenvectors on PASCAL VOC 2012.}
    Examples of images and the first $4$ corresponding eigenvectors of our feature affinity matrix (excluding the zero-th constant eigenvector).
    The eigenvectors correspond to semantic regions, with the first eigenvector usually identifying the most salient object in the image. 
  }\label{fig:eig_examples}
\end{figure}

In \cref{fig:eig_examples}, we show qualitative examples of our eigendecomposition.
We find that the eigensegments correspond to semantically meaningful image regions and have well-delineated boundaries. 
As such, localization and segmentations tasks arise as natural immediate applications of this approach. 
Importantly, if the choice of $\phi$ is a model trained with self-supervision on unlabeled data, then it is possible to address these tasks without supervision and without the need for finetuning.  

\subsection{Object Localization}

Object localization is the task of identifying, in the form of a bounding box, the location of the primary object in an image. 
To localize the main object, we simply follow the standard spectral bisection approach. 
We examine the Fiedler eigenvector $y_1$ of $L$ and discretize it by taking its sign to obtain a binary segmentation of an image. 
We then take a bounding box around the smaller of the two regions, which is more likely to correspond to any foreground object rather than the background. 

\subsection{Object Segmentation}

Object segmentation, also called foreground-background segmentation, is a binary segmentation task that is very closely related to object localization; it consists of densely segmenting the foreground object in an image. We approach this task in the same manner as object localization, by using the Fiedler eigenvector $y_1$ to first find a coarse object segmentation. However, rather than taking a bounding box around our coarse segment, we apply a simple pairwise CRF to increase the resolution of our segmentation from $M'\times N'$ to $M\times N$. 

\subsection{Semantic Segmentation}

Semantic segmentation is the task of categorizing each pixel in an image with a label that is semantically consistent across an entire dataset. Ideally, we would approach this problem by constructing a Laplacian matrix $L$ of size $MNT \times MNT$ containing the pairwise affinities of all pixels in an entire dataset of size $T$, but this is computationally infeasible. As a result, we perform a three-step process in which we: (1) break each image into segments, (2) compute a feature vector for each segment, and (3) cluster these segments across a collection of images. 

For step (1), we discretize the first $m$ eigenvectors $\{y_1, \cdots, y_m\}$ of $L$ by clustering them across the eigenvector dimension using $K$-means clustering (for every image separately). 
Since we do not know a-priori the number of meaningful segments for an image, we find empirically that it is preferable to over-cluster the image into regions.
For step (2), we take a crop around each segment, and compute its feature vector $f_{s}$ using our self-supervised transformer.
For step (3), we cluster the set of all feature vectors $\{ f_{s} \}$ across all images using $K$-means clustering with $k$ clusters. 
The second clustering step assigns a label to each segment of each image; adjacent regions with the same label are merged together, effectively reducing the number of segments found per image as needed.
At the end of this process, we arrive at a set of semantic image segmentations consistent across the entire dataset.

Finally, we apply a self-training step to further improve segmentation quality. 
We train a standard segmentation model with a self-supervised backbone 
using the segmentations obtained above as pseudolabels. 
As shown in \cref{fig:seg_comparison}, this distillation process improves segmentation quality and inter-image consistency.

\begin{figure}
  \centering
   \includegraphics[width=0.47\textwidth]{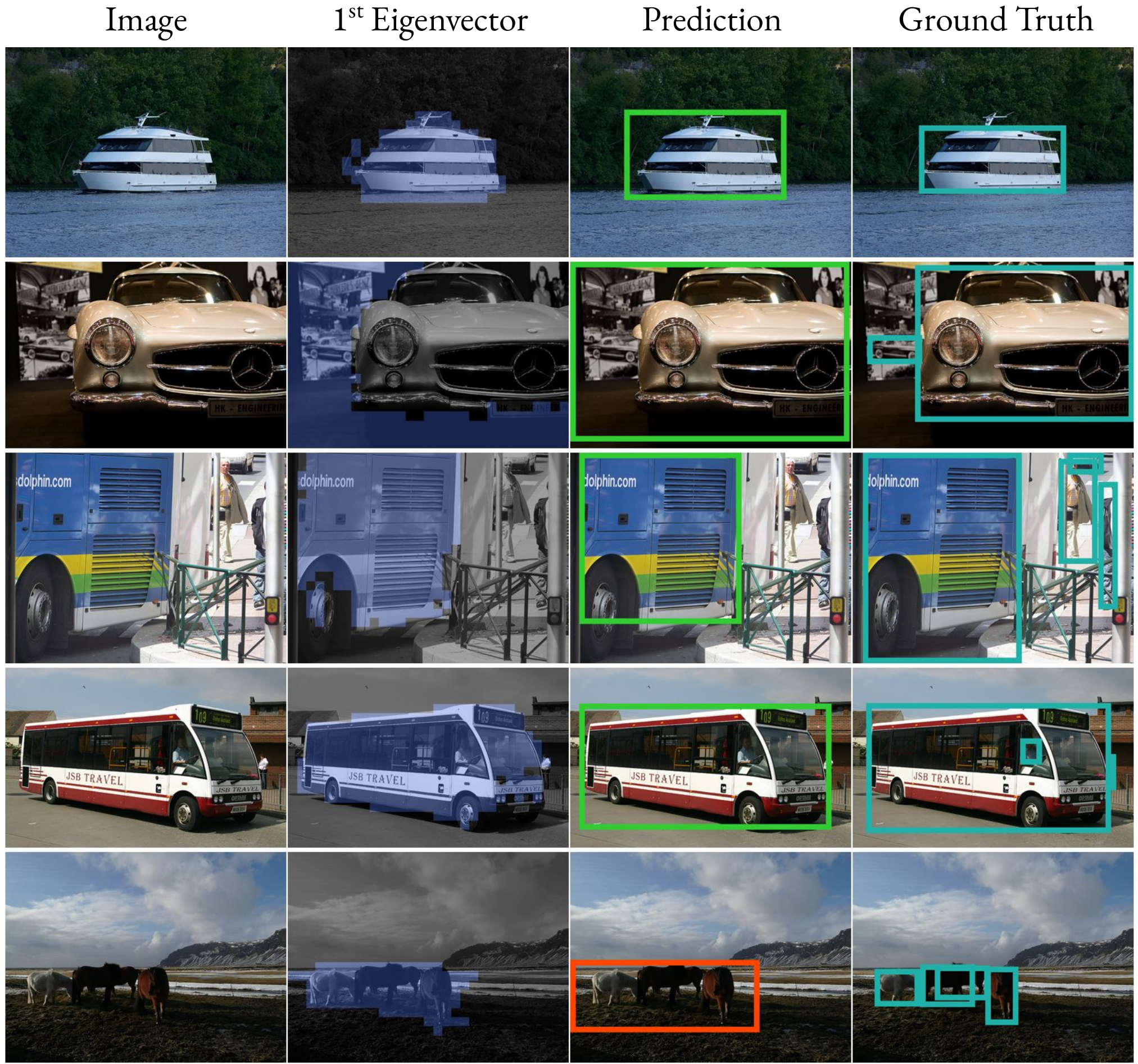}
  \caption{\textbf{Object localization on PASCAL VOC 2012}. From left to right, we show the original image, a visualization of the mask produced by thresholding the Fiedler eigenvector at 0, our predicted bounding boxes, and the ground truth bounding boxes. We color our bounding boxes in green or red based on whether they do or do not have IoU greater than 50\% with one of the ground-truth boxes.
  }
  \label{fig:loc_examples}
\end{figure}

\section{Experiments}\label{s:experiments}

To evaluate the effectiveness of spectral methods for unsupervised tasks, we perform experiments on unsupervised object localization, object segmentation, semantic segmentation and image matting.
We use DINO-ViT-Base~\cite{caron2021emerging} as our self-supervised transformer for feature extraction unless otherwise noted. 
Further implementation details are discussed in the supplementary material.

\subsection{Object Localization}

We compare our unsupervised object localization method to prior work on three benchmarks: PASCAL VOC 2007, PASCAL VOC 2012, and COCO-20k (a subset of 20k images from the MS-COCO dataset~\cite{lin2014microsoft}, introduced in \cite{vo2020toward}). 
We evaluate following the same procedure as \cite{Vo20rOSD,LOST}. 
As is standard practice, we evaluate on the \emph{trainval} sets of these datasets. 
Results are measured in terms of the Correct Localization (CorLoc) metric, which measures the percentage of images on which an object is correctly localized. 
Since the images typically contain multiple objects, a prediction bounding box is considered to have correctly identified an object if it has greater than 50\% intersection-over-union with \textit{any} ground truth bounding box. 

\begin{table}[t]
\small
\centering
\begin{tabular}{lccc}
\toprule
\textbf{Method}  & \textbf{VOC-07} & \textbf{VOC-12} & \textbf{COCO-20k}  \\
\midrule
Selective Search~\cite{Vo20rOSD}                &  18.8          &  20.9          &  16.0 \\
EdgeBoxes~\cite{uijlings2013selective}          &  31.1          &  31.6          &  28.8 \\
Kim et al.~\cite{kim2009unsupervised}           &  43.9          &  46.4          &  35.1 \\
Zhang et al.~\cite{OLMZhang}                    &  46.2          &  50.5          &  34.8 \\
DDT+~\cite{wei2019unsupervised}                 &  50.2          &  53.1          &  38.2 \\
rOSD~\cite{zitnick2014edge}                     &  54.5          &  55.3          &  48.5 \\
LOD~\cite{vo2021large}                          &  53.6          &  55.1          &  48.5 \\
DINO-\texttt{[CLS]}~\cite{caron2021emerging}    &  45.8          &  46.2          &  42.1 \\
LOST~\cite{LOST}                                &  61.9          &  64.0          &  50.7 \\ \midrule
Ours                                            &  \textbf{62.7} &  \textbf{66.4} &  \textbf{52.2} \\
\bottomrule
\end{tabular}
\caption{\textbf{Single-object localization performance (CorLoc).} As is standard practice, we use the trainval sets of PASCAL VOC 2007 and 2012 for evaluation.}
\label{table:loc_main_comparison}
\end{table}

We give quantitative results in \Cref{table:loc_main_comparison}.
Compared to older methods based on identifying co-occurances across image collections, our method delivers dramatically improved performance.  
Relative to the state-of-the-art, LOST~\cite{LOST}, which uses the same self-supervised features, our method still meaningfully outperforms across all datasets. 
In \cref{table:loc_ablation_model} we ablate the influence of the architecture on the localization performance and find that transformer-based models score significantly higher than ResNets, with larger models performing better.
\cref{fig:loc_examples} shows qualitative examples of our method. 

In the supplementary material, we conduct a series of ablation studies to better understand our proposed method. Our ablations generally support the idea that the spatial attention mechanism in vision transformers results in the learning of well-localized intermediate features. Concretely, we find that the most effective features for our tasks are those extracted from the attention keys in the later blocks of vision transformers.

\begin{table}[t]
\small
\centering
\begin{tabular}{llcc}
\toprule
\textbf{Model}  & \textbf{Pretraining}  & \textbf{LOST~\cite{LOST}} & \textbf{Ours} \\
\midrule
ResNet-50     &   DINO     &  \textbf{36.8}  &  26.9  \\
\midrule
ViT-S-16      &   MoCo-v3  &  32.5 &  \textbf{37.3} \\
ViT-S-16      &   DINO     &  \textbf{61.9} &  61.6 \\
\midrule
ViT-B-16      &   MoCo-v3  &  53.3 &  \textbf{61.1} \\
ViT-B-16      &   DINO     &  60.1 &  \textbf{61.6} \\
\midrule
ViT-S-8       &   DINO     &  55.5 &  \textbf{62.6} \\
\midrule
ViT-B-8       &   DINO     &  46.6 &  \underline{\textbf{62.7}} \\
\bottomrule
\end{tabular}
\caption{\textbf{Architecture and pretraining.} Single-object localization performance on PASCAL VOC 2007. All models are trained with $\lambda_\mathrm{knn} = 0$ to isolate the quality of the features. 
Vision transformers outperform the convolutional ResNet-50 model, possibly due to the inherent structure of self-attention layers.%
}
\label{table:loc_ablation_model}
\end{table}

\subsection{Object Segmentation}

For single-object segmentation, we evaluate on four challenging benchmarks: CUB~\cite{wah11the-caltech-ucsd}, DUTS~\cite{wang2017learning}, DUT-OMRON~\cite{yang2013saliency}, and ECSSD~\cite{shi2015hierarchical}. We evaluate in the same manner as \cite{voynov2021object,melas2021finding}, using the standard mean intersection-over-union (mIoU) metric to measure performance. 
Quantitative results are shown in \cref{table:appendix_obj_seg}. Our approach generally outperforms highly-tuned methods based on layerwise GAN learning (\ie, PertGAN~\cite{bielski_emergence_2019}, ReDO~\cite{chen_unsupervised_2019}, OneGAN~\cite{benny_onegan_2020}), extracting segmentations from GANs (\ie \cite{voynov2021object,melas2021finding}), and dense contrastive training (\eg IIC~\cite{ji19invariant}). Qualitative results are shown in the supplement.

\subsection{Semantic Segmentation}

\begin{table}[t]
\small
\centering %
\begin{tabular}{l c c c c c}
\toprule
\textbf{Method}                            &  \textbf{CUB} &  \textbf{DUTS} & \textbf{OMR} &  \textbf{ECSSD} \\ \midrule
{\cite{bielski_emergence_2019}  PertGAN}      &  0.380                            &  -                            &  -                             &  -  \\
{\cite{chen_unsupervised_2019} ReDO}         &  0.426                            &  -                            &  -                             &  -  \\
{\cite{kanezaki_unsupervised_2018} UISB}     &  0.442                            &  -                            &  -                             &  -  \\
{\cite{ji19invariant} IIC}-seg               &  0.365                            &  -                            &  -                             &  -  \\
{\cite{benny_onegan_2020} OneGAN}            &  0.555                            &  -                            &  -                             &  -  \\
{\cite{voynov2021object} Voynov \etal}       &  0.683                            &  0.498                        &  0.453                         &  0.672  \\
{\cite{melas2021finding} {\footnotesize Melas-Kyriazi} \etal} &  0.664                            &  \textbf{0.528}               &  0.509                         &  0.713  \\ \midrule
\textit{Ours}                     &  \textbf{0.769}                   &  0.514                        &  \textbf{0.567}                &  \textbf{0.733}  \\
\bottomrule
\end{tabular}
\caption{\textbf{Single-Object Segmentation.} 
We present mIoU scores across four datasets, comparing to highly-tuned methods based on GANs \cite{bielski_emergence_2019, chen_unsupervised_2019, benny_onegan_2020,voynov2021object,melas2021finding} and contrastive learning \cite{ji19invariant}.
}
\label{table:appendix_obj_seg}
\end{table}

\begin{table}[t]
\small
\centering
\def\arraystretch{0.85}
\begin{tabular}{l@{\hskip 30pt}r}
\toprule
\textbf{Method}  & \textbf{mIoU}  \\
\midrule
\multicolumn{2}{l}{\textit{\footnotesize Pretext task methods}} \\
\midrule
Co-Occurrence~\cite{isola2015learning}      &  4.0 \\
CMP~\cite{zhan2019self}                     &  4.3 \\
Colorization~\cite{zhang2016colorful}       &  4.9 \\
\midrule
\multicolumn{2}{l}{\textit{\footnotesize Clustering/Contrastive methods}} \\
\midrule
IIC~\cite{ji2018invariant}                  &  9.8 \\
MaskContrast$^{\dagger}$~\cite{vangansbeke2020unsupervised}  &  35.0 \\
\midrule
\multicolumn{2}{l}{\textit{\footnotesize Additional baselines}} \\
\midrule
Cluster-Patch                            &  5.3   \\ 
Cluster-Seg                              &  12.1  \\ 
Saliency-DINO-ViT-B$^{\dagger}$          &  30.1  \\ 
MaskContrast-DINO-ViT-B$^{\dagger}$      &  31.2  \\
\midrule
\textit{Ours w/o self-training}          & 30.8 $\pm$ 2.7  \\ 
\textit{Ours}                            & \textbf{37.2} $\pm$ 3.8  \\ 
\bottomrule
\end{tabular}
\caption{\textbf{Semantic Segmentation on PASCAL VOC 2012.} We calculate the average and standard deviation of our method over four random seeds. $^{\dagger}$ indicates methods using saliency networks that were initialized from supervised models.}
\label{table:seg_comparison}
\end{table}

We now consider the challenging setting of unsupervised semantic segmentation, which differs from object segmentation in that it involves identifying multiple semantic masks per image and associating these masks across images. 
We evaluate our approach on the PASCAL VOC 2012 dataset, which contains 21 semantic classes (20 classes and a background class). 
For the segmentation phase, we start by clustering the eigenvectors into 15 segments per image.
We then compute a feature vector for each of these segments, and cluster these across the dataset using $K$-means with $k = 21$ clusters. 
For the self-training stage, we train a DeepLab~\cite{deeplab} model with a ResNet-50~\cite{resnet} backbone that is pre-trained with self-supervision using DINO.
We freeze the first two layers of the ResNet backbone and train the rest using the AdamW~\cite{kingma14adam:} optimizer with learning rate of 0.005 for 3000 steps.
We perform self-training on the training set with a standard cross-entropy loss. 
Finally, we evaluate the outputs of the final self-trained model on the validation set using mIoU. 
Since clustering results in pseudo-labeling of image segments, Hungarian matching~\cite{kuhn1955hungarian} is used to optimally match clusters to ground truth labels. 

In \cref{table:seg_comparison}, we compare to prior methods, including the state-of-the-art MaskContrast~\cite{vangansbeke2020unsupervised}. 
We note that MaskContrast uses saliency masks from Deep-USPS~\cite{nguyen_deepusps_2019}, which was pre-trained with supervision using labeled data (Cityscapes~\cite{cordts2016cityscapes}). 
For a fairer comparison, we also train a version of MaskContrast based on a DINO-pretrained model (MaskContrast-DINO-ViT-B). 
Additionally, we give results for directly clustering DINO-pretrained features masked with Deep-USPS saliency maps (Saliency-DINO-ViT-B), \ie, without the additional dense training; this alone almost reaches the performance of MaskContrast. 

Finally, we compare to two additional clustering-based baselines. 
The first baseline (Cluster-Patch) is to directly cluster the self-supervised image features for all patches across the entire dataset. 
That is, we extract $\frac{MN}{P^2} \cdot T$ features from all $T$ images and assign them to clusters via $K$-means with 21 clusters. 
The second baseline (Cluster-Seg) uses $K$-means (in feature space) for each image individually to obtain class-agnostic segments, then computes the average feature within each segment, and finally clusters segments over the entire dataset (by using $K$-means with $K=21$, as above). 

\begin{figure}[t]
  \centering
  \includegraphics[width=0.47\textwidth]{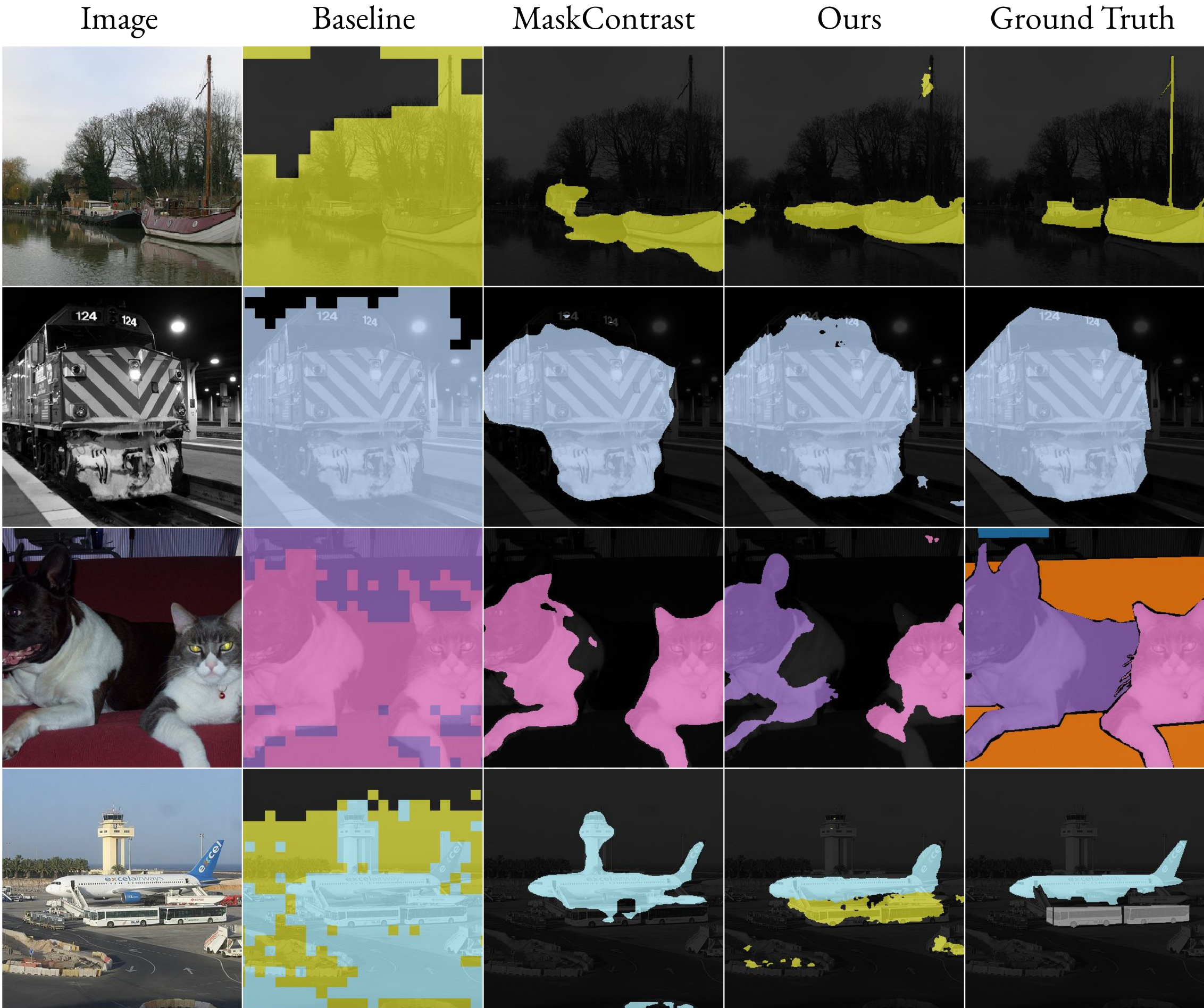}
  \caption{\textbf{Sematic Segmentation on PASCAL VOC 2012.}
    From left to right: we show the original image, the Cluster-Seg baseline, MaskContrast~\cite{vangansbeke2020unsupervised} segmentation, our results and the ground truth.
    The Cluster-Seg baseline often fails to adequately capture object boundaries.
    Whereas MaskContrast is only able to identify a single semantic class per image, our method successfully segments multiple semantic objects in the same image. 
  }\label{fig:seg_comparison}
\end{figure}

Qualitative results are shown in \cref{fig:seg_comparison}. 
Most notably, in contrast to \cite{vangansbeke2020unsupervised}, we are able to segment multiple different semantic regions in the same image. For example, in the third row of the figure, our method correctly segments an image containing both a dog and a cat, whereas MaskContrast is limited to a single label (cat) for both regions.

\subsection{Image Matting}

Finally, we show that with a slight modification, our method can be used for real-world image editing tasks. 
The only modification we make is that we perform the spectral clustering at full resolution rather than at a reduced intermediate resolution.
We sparsify the feature affinity matrix $W_\mathrm{feat}$ by randomly subsampling $\frac{1}{16^2} \approx 0.4\%$ of its entries. 
Note that despite the sparsity of $W$, computing its eigenvalues at full $(MN \times MN)$ resolution is still too slow to be performed on an entire dataset,\footnote{Computing the eigenvectors of $W$ takes around 1 min on 100 CPUs, which corresponds to $\approx$ 10 days for all of PASCAL VOC 2012.} which is why we use a lower resolution for the other experiments. 
However, if we are only concerned with editing a small number of images, then computing at full resolution is perfectly feasible. 

\begin{figure}[t]
  \centering
  \includegraphics[width=0.47\textwidth]{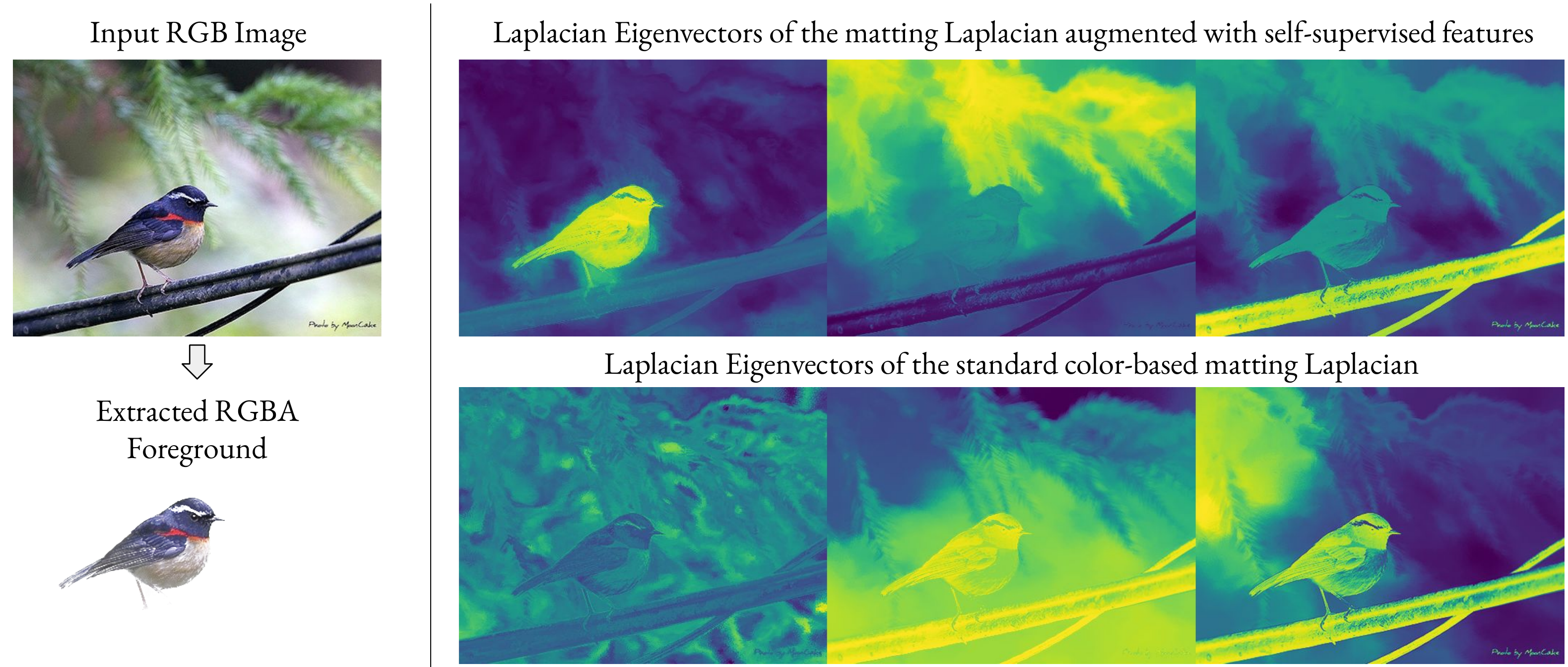}
  \caption{\textbf{Image matting.}
    Combining semantic and color information yields soft image mattes that are well-suited to image editing. 
    Left: the original image and the extracted foreground object using our matting formulation. 
    Right (above): eigensegments derived from our affinity matrix correspond to semantically consistent regions. 
    Right (below): eigensegments from traditional color-based matting do not capture semantic regions well. 
  }\label{fig:matting_examples}
  \vspace{-3mm}
\end{figure}

In \cref{fig:matting_examples}, we show examples of our mattes, with and without the feature affinity term. 
We see that the mattes are significantly more useful for editing with our feature affinity term, as they better correspond to objects in the scene. 
These mattes can then be used as primitives for further computer graphics operations, such as foreground extraction, selective colorization, or background replacement. 

\section{Conclusions}\label{s:conclusions}

In this paper we introduced a method for unsupervised localization, segmentation and matting based on spectral graph theory and deep features. 
Despite the simple formulation, it achieves state-of-the-art unsupervised performance for these tasks. 
It is interesting to note that this performance is only achieved with features from transformer architectures and not with CNNs, which we attribute to the inherent functionality of self-attention in transformers that aligns well with dense localization tasks.
While spectral graph theory has been relegated to a minor role in the age of deep learning, we find that the inductive biases on which it is built can be very useful in the unsupervised setting.

\section*{Acknowledgements}\label{s:acknowledgements}
L.\ M.\ K.\ is supported by the Rhodes Trust. C.\ R.\ is supported by Innovate UK (project 71653) on behalf of UK Research and Innovation (UKRI) and by the Department of Engineering Science at Oxford. I.\ L.\ and A.\ V.\ are supported by the VisualAI EPSRC programme grant (EP/T028572/1). 

\clearpage

\clearpage
{ 
\small
\bibliographystyle{ieee_fullname}
\bibliography{bibliography/new,bibliography/vedaldi_specific,bibliography/vedaldi_general}
}

\end{document}


\title{
    Deep Spectral Methods: A Surprisingly Strong Baseline for Unsupervised Semantic Segmentation and Localization \\ \vspace{3mm} {\large \textit{Supplementary Material}}
}
\maketitle

\setcounter{figure}{6}
\setcounter{table}{5}

\section{Implementation Details} \label{s:appendix}

Here, we present full implementations details for reproducing our results. 

\paragraph{Spectral Decomposition}

We operate on images at their full resolution. For the spectral decomposition step, we start by extracting the key features $f \in \mathbb{R}^{C \times M/P \times N/P}$ from the final layer of the vision transformer. We then normalize these features along the embedding dimension and compute the affinity matrix $W_\mathrm{feat}$.
Next, we calculate color features $W_\mathrm{knn}$ by first downsampling the image to the intermediate resolution resolution $M' \times N'$ and then calculating the sparse KNN affinity matrix using the implementation from PyMatting~\cite{germer2020pymatting}. We use $M' = M/8$ and $N' = N/8$ for all experiments.

Next we perform the fusion of color and feature information. If $P = 8$, then $\frac{MN}{P^2} = M' \cdot N'$, so the feature affinities above are already at the intermediate resolution. If $P = 16$, the feature affinities are upscaled $2 \times$ to size $M' \times N'$. The affinities are summed, weighted by $\lambda_\mathrm{knn}$
and the eigenvectors of the Laplacian $L$ are calculated using the Lanczos algorithm. 

We also need to address the fact that vision transformers only operates on images whose size is a multiple of the patch size $P$. Since we operate on images at their full resolution, which is not always a multiple of $P$, this presents a small complication. To resolve this, we simply crop images to the nearest multiple of $P$ by truncating the right and top edges. For object localization, this means that we do not predict bounding box coordinates in the cropped edge regions. For the segmentation task, we compute segmentations on the cropped image and then simply replicate the right and bottom edges until the segmentation matches the original image. Fortunately, since the patch size is quite small relative to the image resolution, this cropping is not a large issue for the vast majority of images. 

\paragraph{Object Localization} For the object segmentation task, we consider the smallest eigenvector $y_1 \in \mathbb{R}^{\frac{MN}{P^2}}$ of $L$ with nonzero eigenvalue. We reshape $y_1$ to size $\frac{M}{P} \times \frac{N}{P}$ and compute its largest fully-connected component. We then draw a bounding box $(x_1, y_1, x_2, y_2)$ around this component and multiply its coordinates by $P$ to obtain the bounding box in the scale of the original image. 

\paragraph{Unsupervised single-object segmentation}

Single-object segmentation is a natural extension of our localization pipeline.  As above, we first using the Fiedler eigenvector to find a coarse object segmentation. We then apply a pairwise CRF to increase the resolution of our segmentations back to the original image resolution $(M, N)$. For the CRF, we use the implementation from \cite{krahenbuhl11efficient} and leave all parameters on their default settings.

\paragraph{Semantic Segmentation} For the semantic segmentation task, we consider the $n$ smallest eigenvectors $\{y_i: i < K\}$ of $L$, reshaped into a tensor of size $n \times \frac{M}{P} \times \frac{N}{P}$. All experiments use $n = 15$. We cluster these eigenvectors using $k$-means clustering with $k = 15$ to obtain a discrete (non-semantic) segmentation of size $\frac{M}{P} \times \frac{N}{P}$ for each image; this breaks the image into $k$ separate segments/regions. In each image, the largest segment is considered to be the background region. We then compute a feature vector for each of the $(k - 1) \times T$ non-background segments in our dataset of size $T$. This feature vector is computed by taking a bounding box around each segment, expanding this bounding box by $2$ patches, cropping this region of the image, and applying the self-supervised transformer to this cropped image. These segment features are clustered over the dataset via $K$-means clustering with $K=20$ (which is the number of non-background classes in PASCAL VOC). Finally, associating each segment in each image with its cluster produces (low-resolution) semantic segmentations.
We carry out the above steps on the \texttt{train\_aug} set of PASCAL VOC.

For the self-training stage, we begin by upscaling each (low-resolution) semantic segmentation obtained in the previous step to the original image resolution. We train a ResNet-50 with a pretrained DINO~\cite{caron2021emerging} backbone and a DeepLab~\cite{deeplab} head. We train for 2000 steps with the Adam~\cite{kingma14adam:} optimizer, learning rate $1\cdot10^{-4}$, batch size $144$, and random crops of size $224$. We decay the learning rate to zero linearly over the course of training. This setup was not extensively tuned and could likely be improved with a comprehensive hyperparameter sweep.
Finally, we evaluate this model on the \texttt{val} set of PASCAL VOC.

\paragraph{Computational Requirements} All experiments are performed on a single NVidia GPU with $16$GB of memory. The eigenvector computation is performed on the CPU and takes approximately $0.5$s for an image of size $512$px at intermediate resolution $H' = W' = H/8 = W/8 = 32$px. Training the semantic segmentation network from the pseudolabeled images takes approximately $2$ hours to finish $2000$ steps. These low computational requirements are one of the strengths of our method, and make it a sensible baseline for future work.

\section{Additional Qualitative Examples}

\begin{figure*}[t]
\centering
\includegraphics[width=0.95\textwidth]{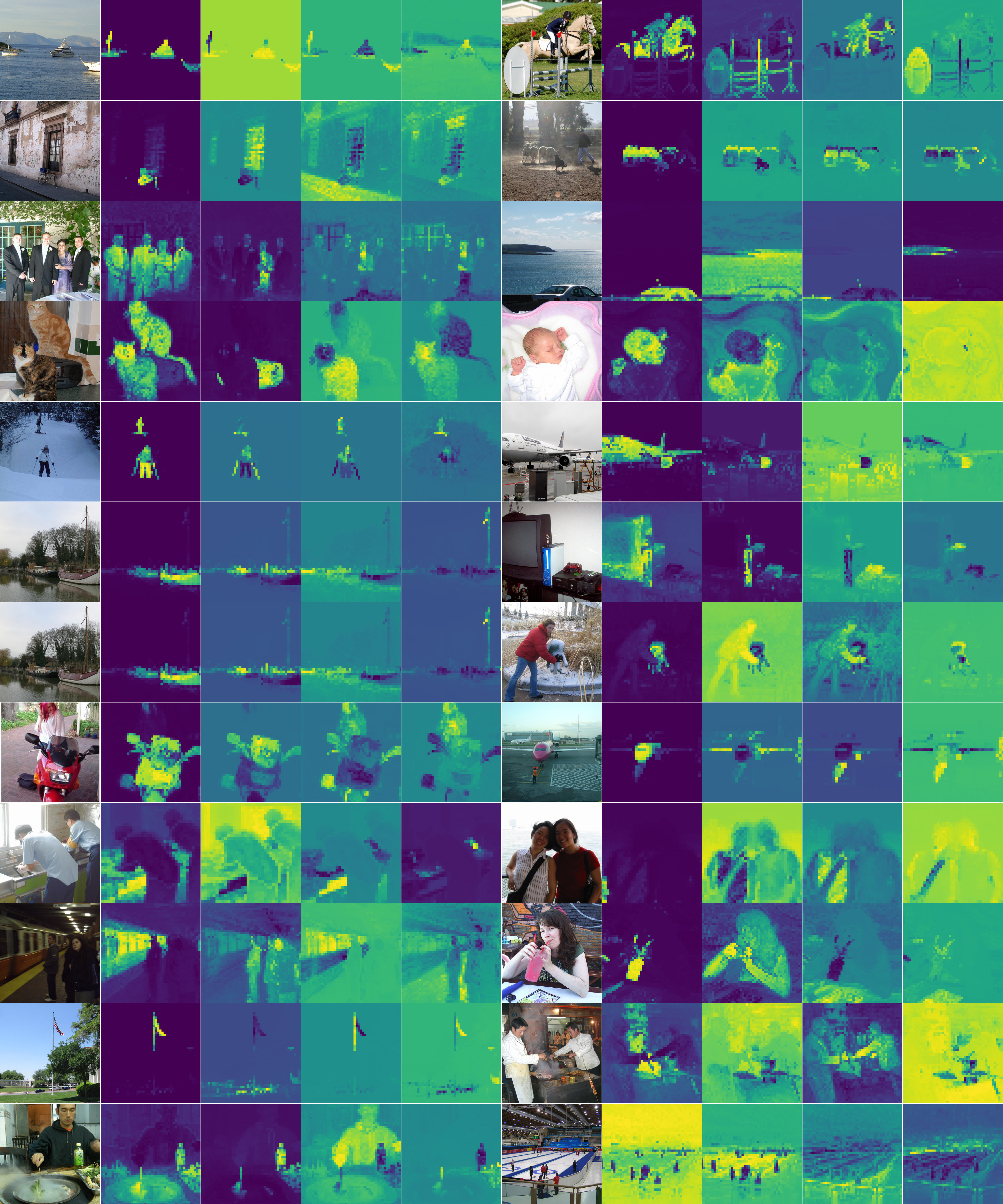}
\caption{
\textbf{Additional examples of eigenvectors extracted by our method on random images from PASCAL VOC 2012.} The first column in each column shows the original image, while the following three columns show the first three eigenvectors.
}
\label{fig:more_examples_eig}
\end{figure*}

\begin{figure*}
\centering
\includegraphics[width=\textwidth]{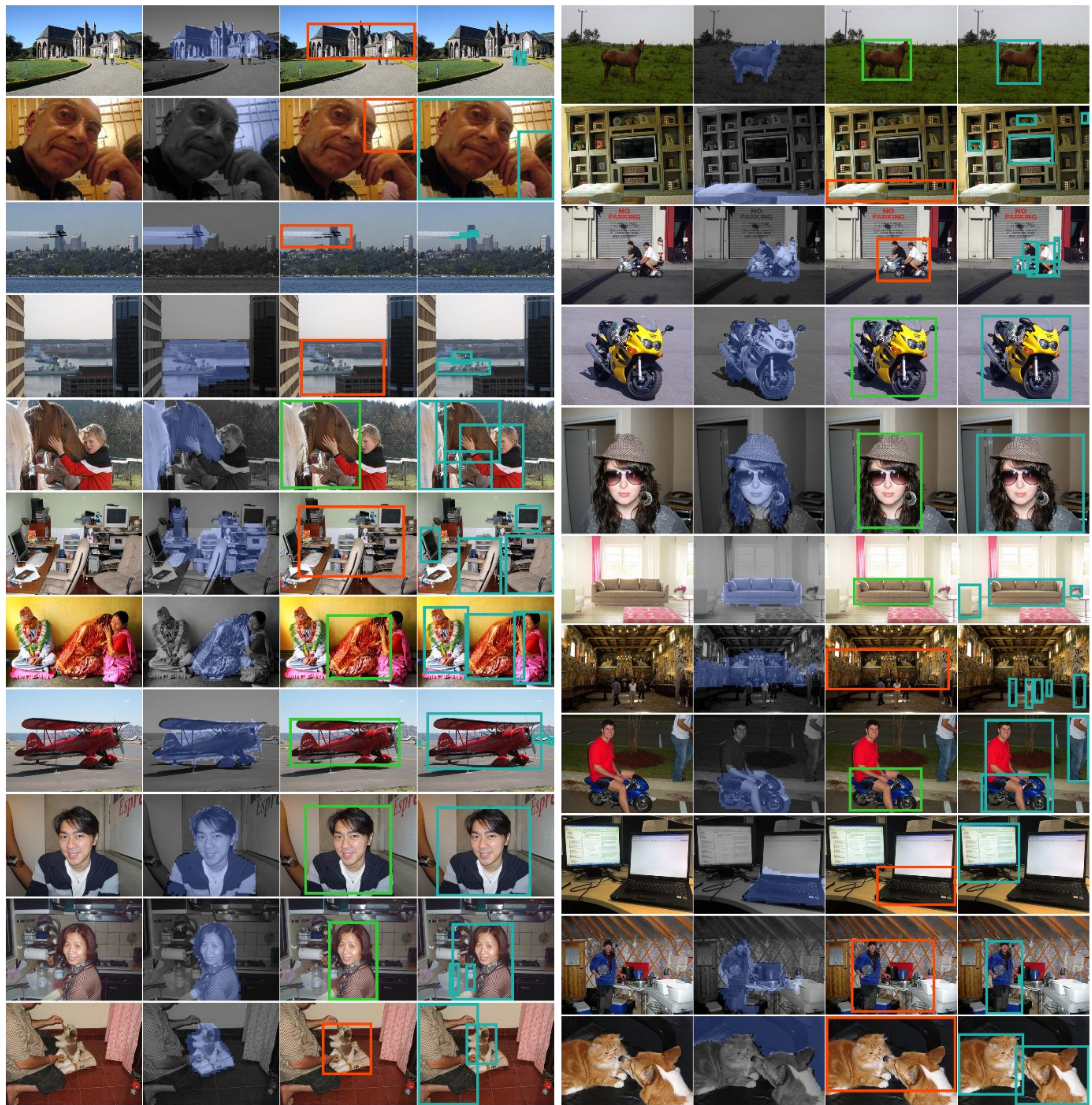}
\caption{
\textbf{Additional object localization examples of our method on random images from PASCAL VOC 2012.}
The first column shows the original image, while the following three columns show our first eigenvector (thresholded at zero), our predicted bounding box, and the ground truth bounding box, respectively. Our bounding box is colored in green or red based on whether it has greater than 50\% mIoU with one of the ground-truth bounding boxes.
}
\label{fig:more_examples_loc}
\end{figure*}

\begin{figure*}
\centering
\includegraphics[width=\textwidth]{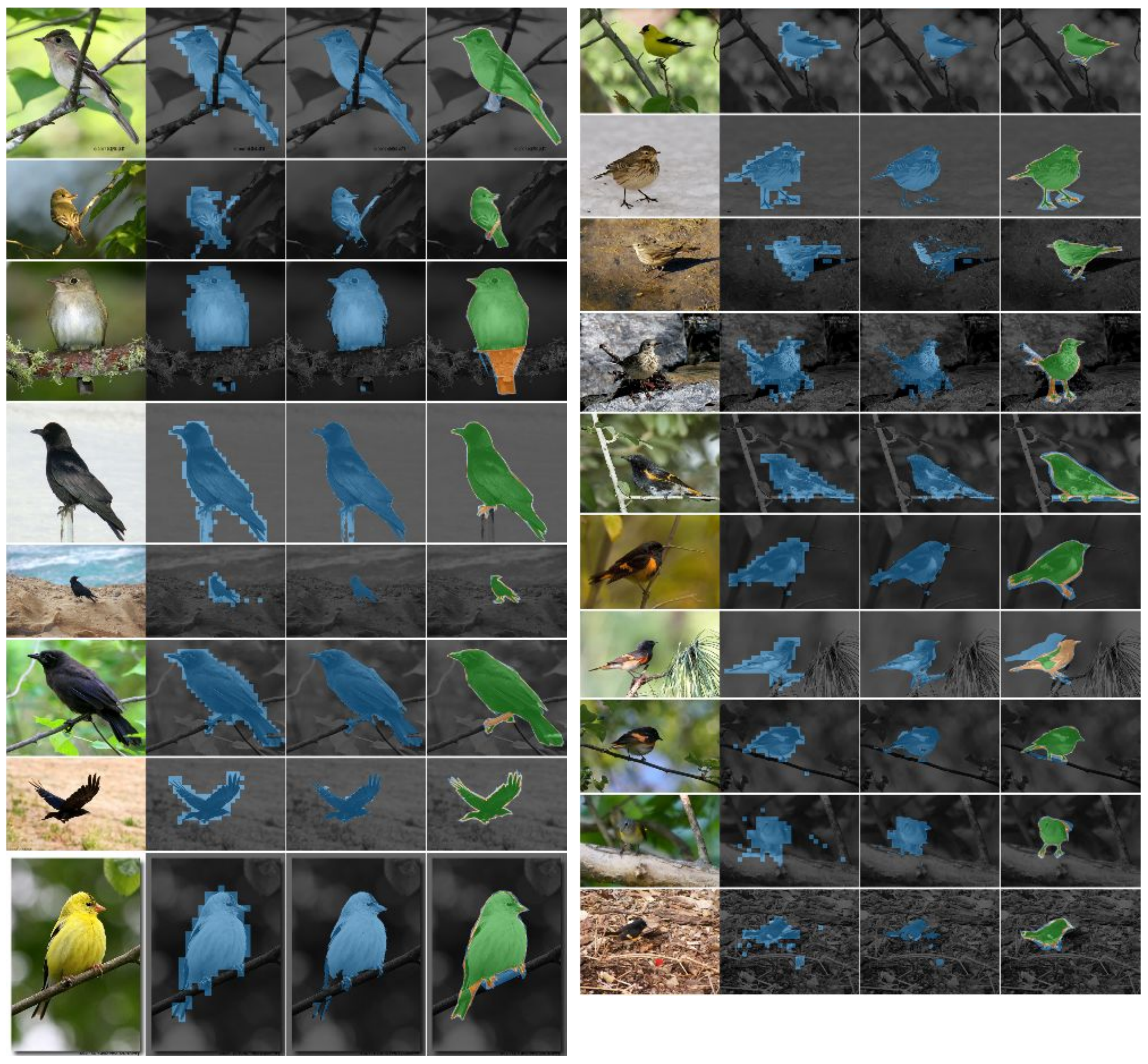}
\caption{
\textbf{Examples of our method for the single-object segmentation task on random images from CUB.}
The first row in each column shows the original image, while the following three rows show our first eigenvector (thresholded at zero), our predicted segmentation, and the ground-truth segmentation. Our segmentation masks accurately locate the bird, often segmenting it without including other objects such as branches or leaves (which is a common failure point of prior state-of-the-art methods). Note that these images are not cherry picked in any way; they are the first images in the CUB dataset.
}
\label{fig:more_examples_obj}
\end{figure*}

\begin{figure*}
\centering
\includegraphics[width=\textwidth]{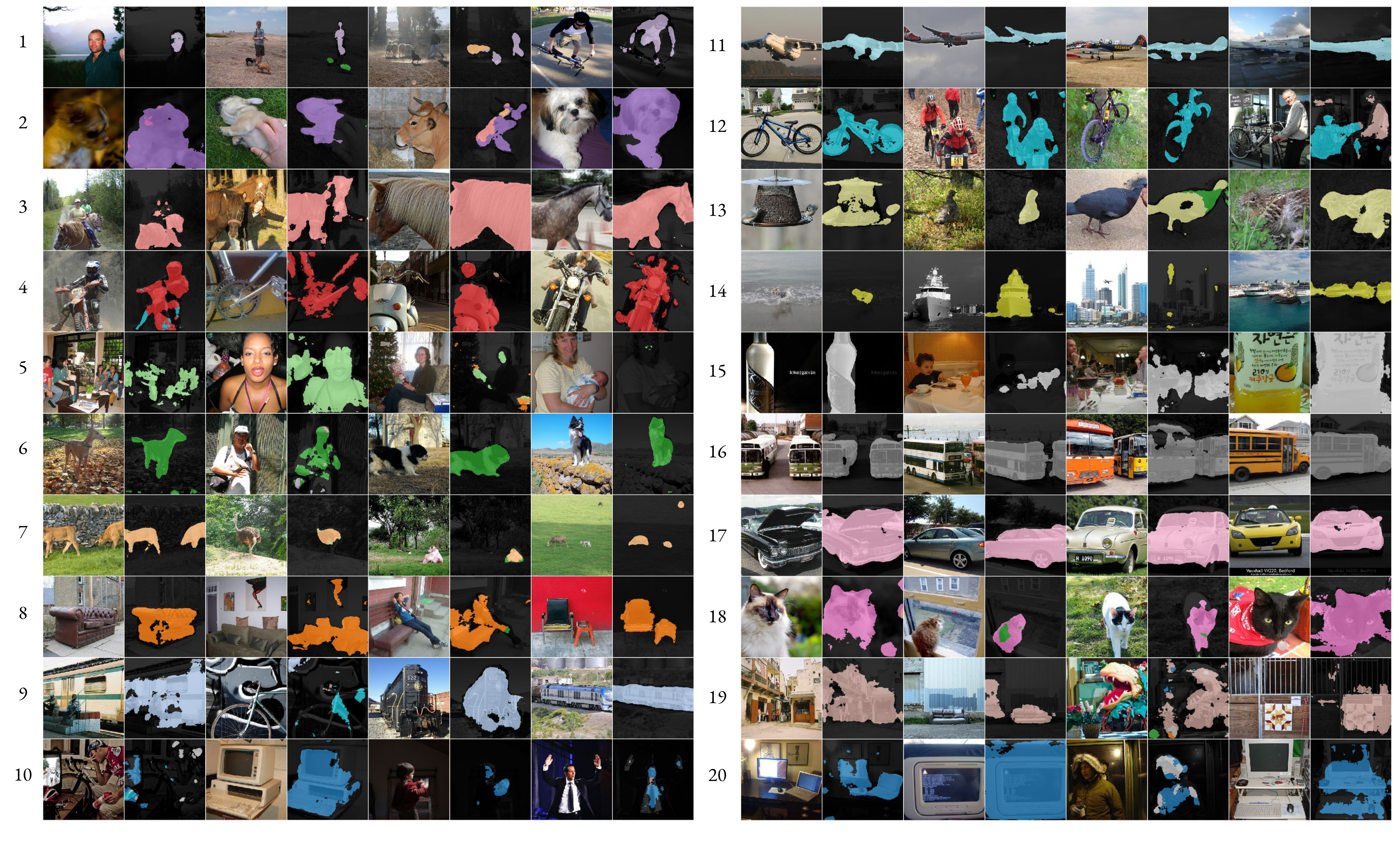}
\caption{
\textbf{Per-pseudoclass examples of our method for the semantic segmentation task on random images from the validation set of PASCAL VOC 2012.}
For each pseudoclass (\ie cluster), we show four randomly selected images from PASCAL VOC for which the given class is the largest segmented region in the object. We see that our pseudoclasses correspond to numerous identifiable concept such as people, buses, boats, cats, airplanes, and bicycles without any human supervision.}
\label{fig:more_examples_semantic_classes}
\end{figure*}

We show additional qualitative examples of our method, including the extracted eigenvectors (\cref{fig:more_examples_eig}) and results on single-object localization (\cref{fig:more_examples_loc}) and single-object segmentation (\cref{fig:more_examples_obj}).
Finally, for the semantic segmentation task, we show the classes discovered by our approach.  
It is important to note that, since our approach is fully unsupervised and uses self-supervised features, eigensegment clusters do not necessarily align with the semantic categories annotated in the dataset (in this case, PASCAL VOC).
In \cref{fig:more_examples_semantic_classes} we show the $K=21$ categories found in the DINO feature space by our method and observe that, while some align with the ground truth label set (e.g., bus, dog, airplane, train), others do not (e.g., cluster 2 seems to be `animal heads` instead of a species-specific cluster). 

\section{Additional Experiments and Ablations}

In this section, we perform additional experiments and ablations in order to better understand the strengths and weaknesses of our method. 

\paragraph{Ablation: Color Information}
To understand the importance of adding color information to the features we evaluate the performance for different values of $\lambda_\mathrm{knn}$ in \cref{table:loc_ablation_lambda_color}. When $\lambda_\mathrm{knn} = 0$, the normalized Laplacian matrix $L$ is composed purely of semantic information. When $\lambda_\mathrm{knn} \to \infty$, the normalized Laplacian matrix $L$ is composed purely of color information.
Larger models and variants with smaller patches benefit less from additional color information as their features likely inherently contain more local image information. Using color features alone is ineffective.

\paragraph{Ablation: Feature Type} In \cref{table:loc_ablation_feature}, we present results when features are extracted from different parts of a self-attention layer. Features from attention \emph{keys} perform best by a large margin, in line with the intuition that the key projection layer should align keys into a shared space for subsequent comparison with query vectors.

\paragraph{Ablation: Feature Depth} In \cref{table:rebuttal_block_ablation}, we present results when features are extracted from different blocks of a ViT network. We find that features from later blocks perform better; they contain semantic information which is both spatially-localized and easy to extract with spectral methods.

\paragraph{Ablation: Network Architecture} In \cref{table:rebuttal_arch_ablation}, we present results when features are extracted from three new architectures: ResNet-50, ConvNext~\cite{liu2022convnet}, and XCit~\cite{el-nouby2021xcit}. ConvNext is a purely-convolutional network pretrained (in a supervised manner) on ImageNet, and it may be thought of as an improved ResNet. XCiT is a transformer with cross-covariance attention (i.e. attention over features rather than spatial locations) pretrained using DINO. ConvNext substantially outperforms ResNet-50 under this setup, demonstrating a link between classification performance and unsupervised object localization performance. Nonetheless, both ConvNext and XCiT lag behind the strong performance of the standard ViT, suggesting that the well-localized nature of \textit{spatial self-attention} is one of the keys to the success of our deep spectral segmentation approach.

\paragraph{Ablation: Semantic Segmentation Pipeline} In \cref{table:rebuttal_eig_semseg_ablation} and \cref{table:rebuttal_k_semseg_ablation}, we present ablation results for varying two aspects of our semantic segmentation pipeline: the number of eigenvectors used in the first stage, and the number of clusters $K$ used in the second stage. We see that our method is fairly robust to the number of eigenvectors used in the first stage, unless one uses a very small number of eigenvectors (i.e. fewer than three). 
For the number of clusters $K$, we first note that the value used in the main paper is not chosen empirically, but rather set to the number of classes (incl.\ background) in the PASCAL VOC dataset, as is common for evaluation purposes. 
In the ablation, for $K > 20$ (i.e.  over-clustering), we compute the optimal matching between our predictions and the ground-truth classes.
Thus, larger $K$ yields superior mIoU scores.

\paragraph{Additional Experiment: Class-Agnostic Detection} In \cref{table:rebuttal_detection_ablation}, we give results for a slightly modified detection setting previously explored in \cite{LOST}. In this setting, denoted class-agnostic detection, we train a class-agnostic object detection model by using the bounding boxes obtained from our method as pseudo-labels. For fair comparison, we follow the same training and evaluation procedure as LOST, including all hyperparameters. We see that our method outperforms LOST despite not tuning any hyperparameters for this task. 

\begin{table}[t]
\small
\centering
\begin{tabular}{lr}
\toprule
\textbf{Feature}   & \textbf{CorLoc} \\
\midrule
Final attention key ($k$)     &  \textbf{61.6} \\
Final attention query ($q$)   &  33.1 \\
Final attention value ($v$)   &  49.9 \\
Final attention output ($o$)  &  37.3 \\
\bottomrule
\end{tabular}
\caption{\textbf{Feature type ablation.}
  Single-object localization performance using different features of a DINO-pretrained ViT-S model, evaluated on PASCAL VOC 2007. Key features perform slightly better than value and much better than query or output features.
}
\label{table:loc_ablation_feature}
\vspace{-2mm}
\end{table}

\begin{table}[t]
\small
\centering
\setlength{\tabcolsep}{5pt}
\begin{tabular}{l@{\hskip 2pt}|lcccc}
\toprule
$\lambda_\mathrm{knn}$  &  \textbf{0.0}  &  \textbf{1.0}  &  \textbf{8.0}                    &  \textbf{10.0}          &  \textbf{inf.}   \\
\midrule
ViT-S-16\hspace{3mm} &  58.0 &  60.1 &  \textbf{61.9}          &  61.5          &  -      \\
ViT-B-16\hspace{3mm} &  57.7 &  59.5 &  61.1                   &  \textbf{61.2} &  24.2   \\
ViT-S-8\hspace{3mm}  &  59.4 &  60.0 &  \textbf{62.6}          &  62.5          &  -      \\
ViT-B-8\hspace{3mm}  &  60.5 &  61.2 &  \underline{\textbf{62.7}} &  62.5          &  -      \\
\bottomrule
\end{tabular}
\caption{\textbf{Importance of color information.}
  An ablation of single-object model localization performance for different values of $\lambda_\mathrm{knn}$ on PASCAL VOC 2012. We find that larger models benefit less from color information, similar to models with smaller patches, as their features likely contain more color information per se.
  }
\label{table:loc_ablation_lambda_color}
\end{table}

\begin{table}[t]
\small
\centering
\begin{tabular}{lc}
\toprule
\textbf{Block}  & \textbf{mIoU} \\
\midrule
12 & \textbf{61.6} \\
11 & 61.4 \\
 8 & 50.5 \\
 4 & 28.2 \\
\bottomrule
\end{tabular}
\caption{\textbf{Ablation across ViT blocks.} Single-object localization performance (CorLoc) on PASCAL VOC 2007 using features extracted from different blocks of a ViT-s16 (DINO) model. Note that the model has 12 blocks, so 12 refers to the last block.}
\label{table:rebuttal_block_ablation}
\end{table}

\begin{table}[t]
\small
\centering
\begin{tabular}{lccc}
\toprule
\textbf{Arch.}  & \textbf{Feature} & \textbf{CorLoc} \\
\midrule
ConvNext & Last conv. in stage 2 & \textbf{41.8} \\
ConvNext & Last conv. in stage 3 & 40.7 \\
ConvNext & Last block in stage 2 & 38.8 \\
ConvNext & Last block in stage 3 & 31.2 \\
XCiT & Cross-covariance attention key & 33.6 \\
ResNet-50 & Last block & 26.6 \\
\bottomrule
\end{tabular}
\caption{\textbf{Ablation across new architectures.} Single-object localization performance (CorLoc) on PASCAL VOC 2007 using features extracted from ConvNext~\cite{liu2022convnet} and XCit~\cite{el-nouby2021xcit}.}
\label{table:rebuttal_arch_ablation}
\end{table}

\begin{table}[t]
\small
\centering
\begin{tabular}{lcc}
\toprule
\textbf{Num. Eigs} & \textbf{mIoU} & \textbf{(w/ ST) mIoU}\\
\midrule
 3 & 29.7 & 32.7 \\
 5 & \textbf{33.3} & 36.3 \\
10 & 31.4 & \textbf{37.5} \\
15 & 31.8 & 36.0 \\
\bottomrule
\end{tabular}
\caption{\textbf{Ablation: Number of eigenvectors for semantic segmentation.} We vary the number of eigenvectors $m$ used in the first step of our semantic segmentation pipeline. Using only three eigenvectors performs poorly, as they are not always sufficient to differentiate different segments of complex scenes. Above three eigenvectors, our method is not very sensitive to the exact number of eigenvectors used.}
\label{table:rebuttal_eig_semseg_ablation}
\end{table}

\begin{table}[t]
\small
\centering
\begin{tabular}{lc}
\toprule
\textbf{$K$} & \textbf{mIoU} \\
\midrule
20 & 33.3 \\
30 & 38.5 \\
40 & 42.8 \\
\bottomrule
\end{tabular}
\caption{\textbf{Ablation: Value of K for semantic segmentation.} We vary the number of clusters $K$ used in the second step of our semantic segmentation pipeline. For $K > 20$ (i.e.  over-clustering), we compute the optimal matching between our semantic clusters and the ground-truth classes. Note that, as a result, larger $K$ yield superior mIoU numbers. We report results without self-training.}
\label{table:rebuttal_k_semseg_ablation}
\end{table}

\begin{table}[t]
  \small
  \centering
  \begin{tabular}{lc}
  \toprule
  \textbf{Method}  & \textbf{CorLoc} \\
  \midrule
  LOST (w/ self-training) & 64.5 \\
  Ours (w/ self-training) & \textbf{65.1} \\
  \bottomrule
  \end{tabular}
  \caption{\textbf{Additional experiment: class-agnostic detection on VOC2007.} In this experiment, we train a class-agnostic object detection model by using the bounding boxes obtained from our method as pseudo-labels. For fair comparison, we follow the same training and evaluation procedure as LOST, which refers to this setup as CAD (class-agnostic detection).}
  \label{table:rebuttal_detection_ablation}
  \end{table}

\section{Discussion of Failure Cases}

Here, we show examples of failure cases and discuss their potential causes, with the goal of fascilitating future research into unsupervised segmentation. 

\paragraph{Spectral Decomposition}

\begin{figure*}[t]
  \centering
  \includegraphics[width=0.54\textwidth]{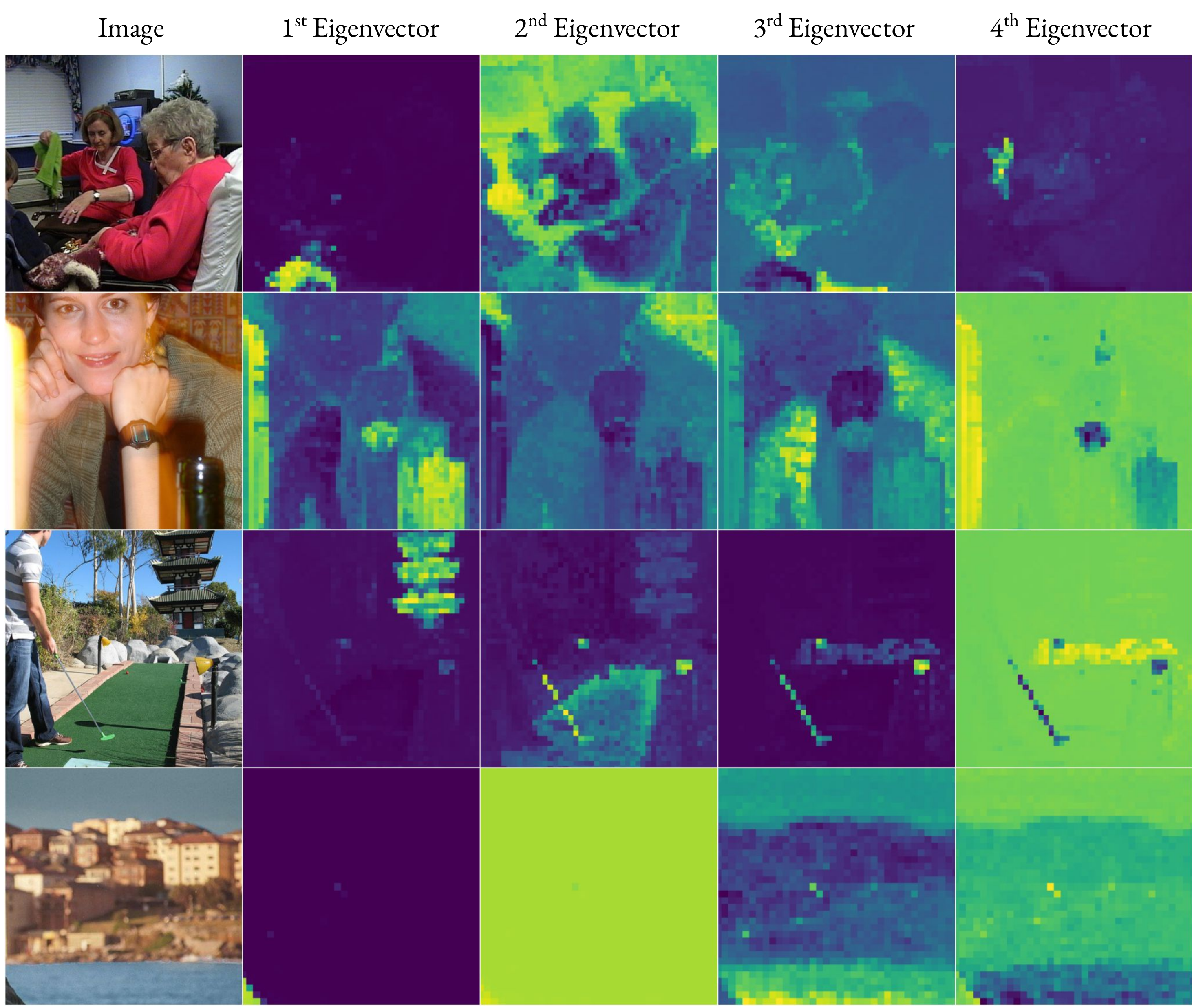}
  \caption{
    \textbf{Examples of failure cases for the eigensegments.} The eigenvectors of the feature Laplacian do not correspond to the primary objects and regions in the scene. These failure cases often occur when a very small object in the foreground lies in the plane of the image, for example in the last image above.
  }\label{fig:eig_failure_cases}
\end{figure*}

Although the notion of a failure case for eigenvectors is not well-defined, we will characterize a failure case as one in which the vectors produced by our method do not match up with our human intuition about the major objects in the scene. We show examples in \cref{fig:eig_failure_cases}. These failure cases often occur when a very small object in the foreground lies in the plane of the image, for example in the last row of the figure. In these cases, the first eigensegment will usually segment this small region. Another failure case in PASCAL VOC occurs when images have borders or frames (these images are present due to the web-scraped nature of the dataset). In these cases, the model nearly always identifies the frame in its first eigenvector.

\paragraph{Object Localization}

We show examples of failure cases for the object localization task in \cref{fig:eig_failure_cases}. When our spectral segmentation method fails, it is usually the result of locating a group of semantically related objects (\eg a group of people) rather than a single entity (\eg an individual person). We note, however, that in many cases these instances are indeed separated by the latter eigenvalues (see Fig. 3 in the main paper); utilizing this information to separate object instances could be an interesting avenue for future research.

\paragraph{Semantic Segmentation}

We show examples of failure cases for the semantic segmentation task in \cref{fig:eig_failure_cases}. We see that the network sometimes fails to detect multiple distinct semantic regions in the same image. Qualitatively, we have observed that this failure mode is actually more common after self-training. In other words, self-training seems to improve performance overall by improving the quality of individual masks, but also seems to hurts the models' ability to segment multiple regions in the same image. There are also some failure cases in which our network should have sharper object boundaries, as is the case with most segmentation networks. 

\section{Additional Related Work}

Here, we discuss relevant work that could not be included in the main paper due to space constraints. 

\paragraph{Unsupervised Segmentation}

Current methods for unsupervised semantic segmentation can broadly be characterized as either generative or discriminative approaches.

For single-object segmentation, generative methods currently rank as the most active research direction, with numerous works having been proposed in the last two years \cite{bielski_emergence_2019,chen_unsupervised_2019,xia2017wnet,kanezaki_unsupervised_2018,ji19invariant,benny_onegan_2020,voynov20big-gans,melas2021finding}. Most commonly, these methods work by generating images in a layerwise fashion and compositing the results. For example, ReDO~\cite{chen_unsupervised_2019} uses a GAN to re-draw new objects on top of existing objects/regions, and Copy-Paste GAN~\cite{arandjelovic2019object} copies part of one image onto the other. 
Labels4Free~\cite{abdal2021labels4free} trains a StyleGAN to generate images in a layer-wise fashion, from which a segmentation may easily be extracted. 
\cite{voynov20big-gans,melas2021finding} extract segmentations from pretrained generative models such as BigBiGAN. 
However, these generative approaches are severely limited in that they only perform foreground-background segmentation: they can only segment a \emph{single} object in each image, and most involve training new GANs. 
As a result, they are not well-suited to segmenting complex scenes nor to assigning semantic labels to objects. 
Another family of methods, most of which adopt a variational approach~\cite{KingmaW13}, focus on unsupervised scene decomposition, effectively segmenting multiple objects in an image~\cite{burgess2019monet, engelcke2019genesis,greff2019multiobject,locatello2020object,emami2021Efficient,lin2020space,crawford2019spatially,monnier2021unsupervised}.
However, these methods cannot assign semantic categories to objects and struggle significantly on complex real-world data~\cite{greff2019multiobject,karazija2021clevrtex}.

Discriminative approaches are primarily based on clustering and contrastive learning. 
Invariant Information Clustering (IIC)~\cite{ji2018invariant} predicts pixel-wise class assignments and maximizes the mutual information between different views of the same image. 
SegSort~\cite{hwang2019segsort} maximizes within-segment similarity and minimizes cross-segment similarity by sorting and clustering pixel embeddings. 
Hierarchical Grouping~\cite{zhang2020self} performs contour detection, recursively merges segments, and then performs contrastive learning. 
MaskContrast~\cite{vangansbeke2020unsupervised}, the current state of the art in unsupervised semantic segmentation, uses saliency detection to find object segments (\ie the foreground) and then learns pixel-wise embeddings via a contrastive objective. 
However, MaskContrast relies heavily on a saliency network which is initialized with a pretrained (fully-supervised) network. 
However, it relies on the assumption that all foreground pixels belong to the same object category, which is not necessarily the case. 
As a result, it is limited to predicting a single class per image without further finetuning.
Finally, DFF~\cite{collins2018deep} uses pre-trained features and performs non-negative matrix factorization for co-segmentation.

\section{Further Description of the Laplacian}

In this section, we present a slightly extended description of the spectral graph theoretic methods that underly our paper. 

Consider a connected, undirected graph $G = (V,E)$ with edges $E$ and vertices $V$. 
We denote by $W = (w_{ij})$ the edge weights between vertices $i$ and $j$. 
In our case, $G$ corresponds to an image $I$, where the vertices $V$ are image patches and the edge weights $W$ are defined by the semantic affinities of patches. 

Let $f: V \to \mathbb{R}$ be a real-valued function defined on the vertices of $G$.
Note that these functions are synonymous with vectors of length $V$, and are thus also synonymous with segmentation maps. 
We begin by asking the question: what does it mean for a function $f$ to be \textit{smooth} with respect to the graph $G$? 

Intuitively, $f$ is smooth when its value at a vertex is similar to its value at each of the vertex's neighbors. 
If we quantify this similarity using the sum of squared errors, we obtain:
\begin{equation} \label{eq:graph_quadratic_form}
  \sum_{(i,j)\in E} (f(i) - f(j))^2 
\end{equation}
which is a symmetric quadratic form. This means that there exists a symmetric matrix $\lapg$ such that 
\[ x^T \lapg x = \sum_{(i,j)\in E} (x_i - x_j)^2 \]
for $x \in \mathbb{R}^n$ with $n = |V|$. This matrix $\lapg$ is the called Laplacian of $G$. 

While there are many ways to define $\lapg$, we present this definition because we believe it gives the greatest insight into the success of our method. 
The standard way of defining $\lapg$ is by the formula $\lapg = D - W$, where $D_{ii} = \deg(i)$ is the diagonal matrix of row-wise sums of $W$. 
The quadratic form definition makes clear some of the fundamental properties of the Laplacian: $\lapg$ is symmetric and positive semi-definite, since $x^T \lapg x \ge 0$ for any $x$.
Additionally, its the smallest eigenvalue is $0$, corresponding to a (non-zero) constant eigenfunction.

To gain intuition for the Laplacian, we present a very simple example from the domain of physics, adapted from \cite{melaskyriazi2020mathematical}. Consider modeling a fluid which flows between a set of reservoirs (vertices) through pipes (edges) with different capacities. Physically, the fluid flow through an edge is proportional to the difference in pressure between its vertices, $x_i - x_j$. Since the total flow into each vertex equals the total flow out, the sum of the flows along a vertex $i$ is $0$: 
\begin{align*}
0 
&= \sum_{j \in N(i) } x_j - \sum_{ j \in N(i)} x_i = \text{deg}(i) x_i - \sum_{j \in N(i)} x_j  \\
&= ((D - W)x)_i  = (\lapg x)_i
\end{align*}
This is known as the Laplace equation $\lapg x = 0$, and it is the simplest special case of Poisson's equation $\lapg x = h$.

The Laplacian spectrum is the centerpiece of our method. 
As described in the main paper, the eigenfunctions of $\lapg$ are orthonormal and form a basis for the space of bounded functions on $G$. 
The Laplacian spectrum does not fully determine the underlying graph, but it nonetheless contains a plethora of information about its structure. 
Our paper leverages this information for a variety of unsupervised dense computer vision tasks.

For further reading in spectral graph theory, we encourage the reader to look into the following resources: \cite{spielman2007spectral,chung1996lectures,spielman2012spectral,melaskyriazi2020mathematical}.

\section{Broader Impact}
It is important to discuss the broader impact of our work with respect to methodological and ethical considerations.

From an ethical perspective, models trained on large-scale datasets\,---\,even the ones considered in our work which are trained without supervision\,---\,might reflect biases and stereotypes introduced during the image collection process~\cite{prabhu2020large,shankar2017no,steed2021image,stock2018convnets}.
In addition, datasets such as ImageNet (used to train the models) and PASCAL VOC (used in our evaluations) contain images from the web (e.g., Flickr).
This data is collected without consent and might also contain inappropriate content~\cite{prabhu2020large}, which raises ethical and legal issues. 
Our method discovers concepts existing in the data through the lens of self-supervised pre-training and, as such, it may be implicitly affected by underlying biases.
For this reason, our method should only be used for research purposes and not in any critical or production applications.

From a methodological perspective, our approach reflects the degree to which different object categories are encoded in the feature space of self-supervised learners.  
Since we do not fine-tune models on a specific dataset for a specific task (e.g., semantic segmentation), these predicted categories may differ from the pre-defined set of categories which are (somewhat arbitrarily) annotated in a given benchmark. 
For example, the categories found by the decomposition and clustering of DINO features, might not necessarily align with the ones annotated in PASCAL VOC; in fact, there is little reason why that should be besides some commonly occurring objects. 
Therefore, how to properly and fairly evaluate fully unsupervised algorithms remains an open question. 

\clearpage

\begin{figure*}
  \centering
  \includegraphics[width=0.55\textwidth]{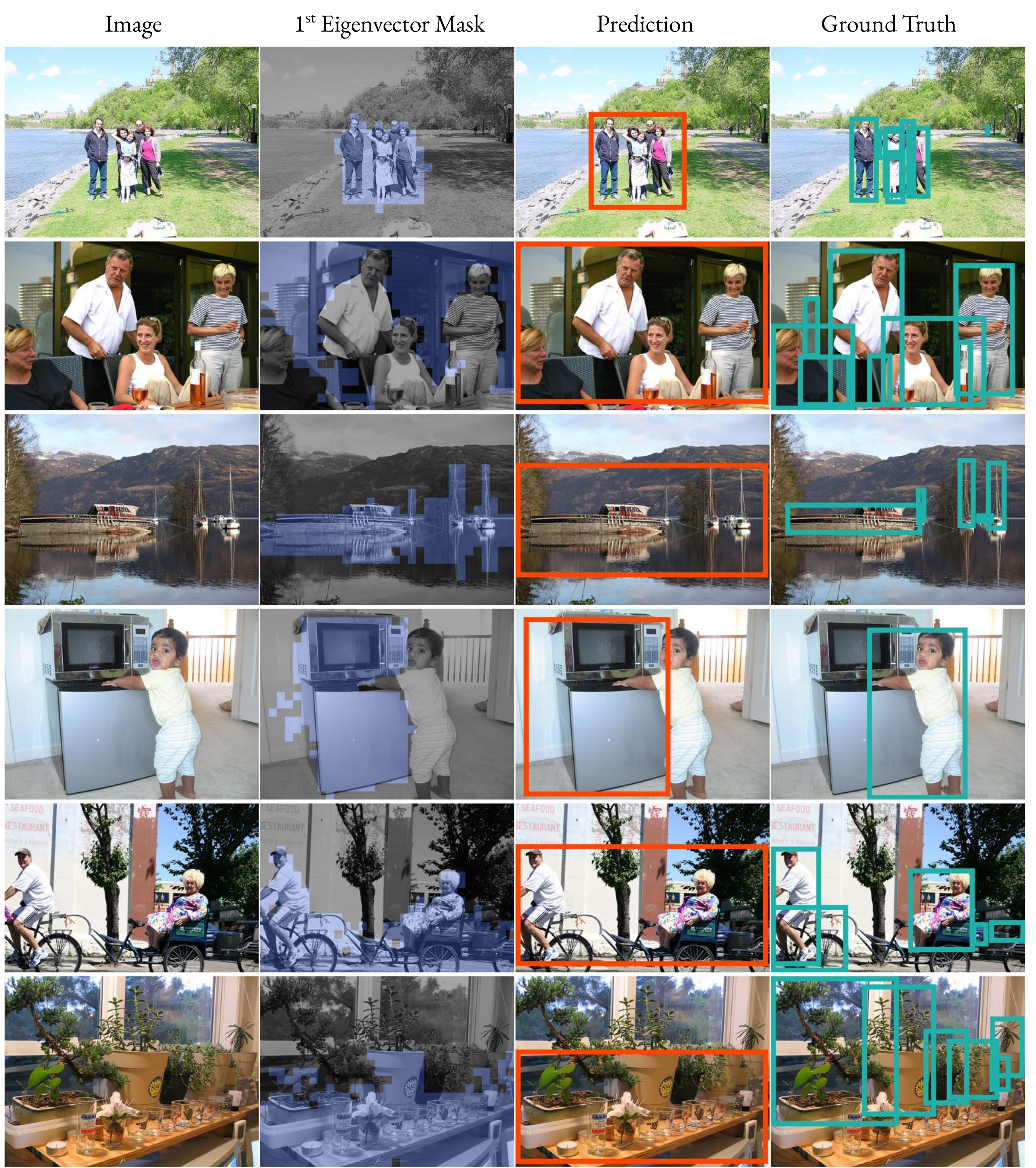}
  \caption{
    \textbf{Examples of failure cases for the object localization task.} When our spectral segmentation method fails, it is usually the result of locating a group of semantically related objects (\eg a group of people) rather than a single entity (\eg an individual person). We note, however, that in many cases these instances are indeed separated by the latter eigenvalues (see Fig. 1); utilizing this information to separate object instances could be an interesting avenue for future research.
  }\label{fig:loc_failure_cases}
\end{figure*}
\begin{figure*}
  \centering
  \includegraphics[width=0.8\textwidth]{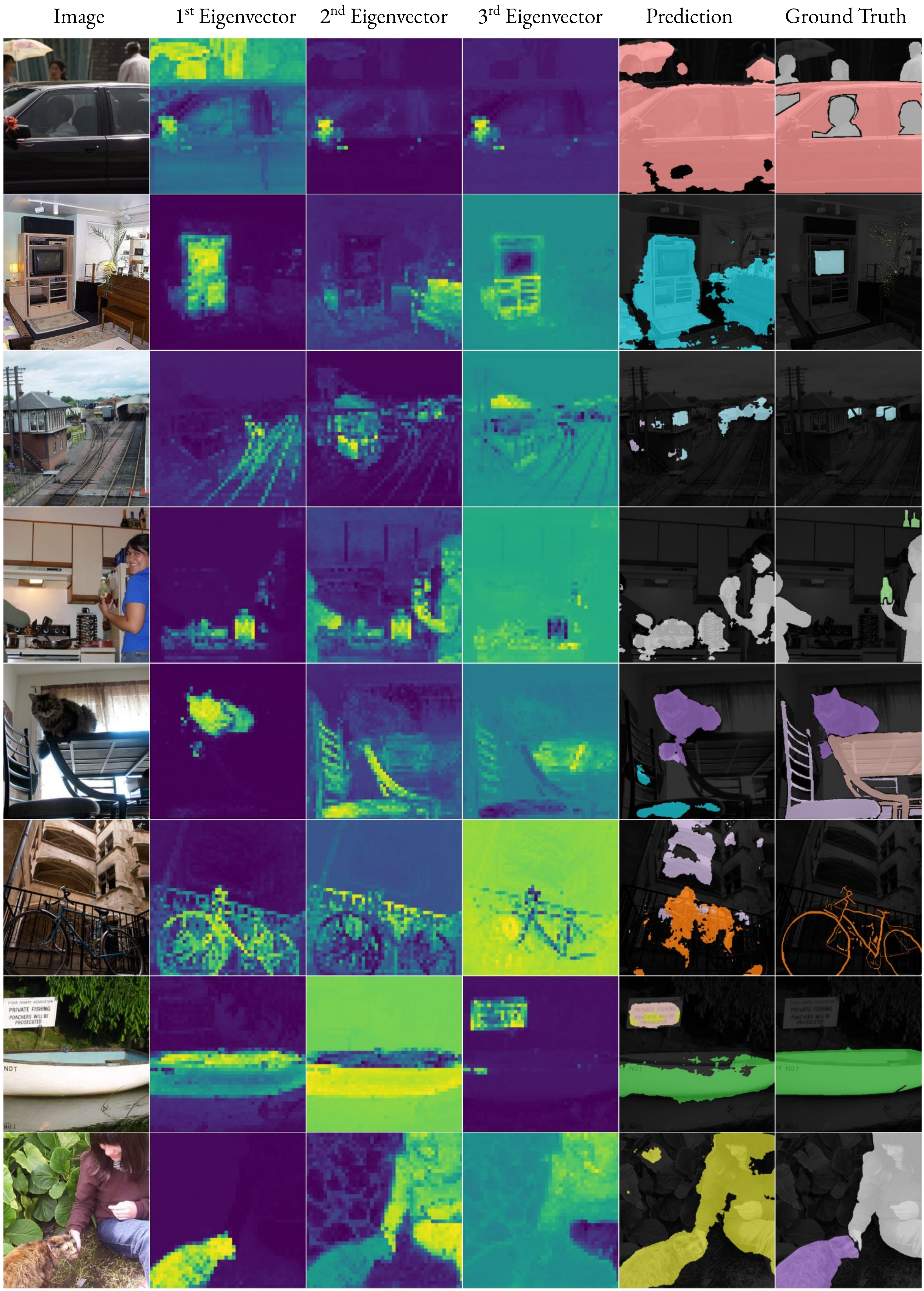}
  \caption{
    \textbf{Examples of failure cases for the semantic segmentation task.} The network sometimes fails to detect multiple distinct semantic regions in the same image. Qualitatively, we have observed that this failure mode is actually more common after self-training. In other words, self-training seems to improve performance overall by improving the quality of individual masks, but also seems to hurts the models' ability to segment multiple regions in the same image. There are also some failure cases in which our network should have sharper object boundaries, as is the case with most segmentation networks. 
  }\label{fig:seg_more_examples}
\end{figure*}

\clearpage

\clearpage

{ 
\small
\bibliographystyle{ieee_fullname}
\bibliography{bibliography/new,bibliography/vedaldi_specific,bibliography/vedaldi_general}
}